\documentclass{article}

\newif\ifanonymous

\usepackage[preprint]{tmlr}
\anonymousfalse

\newcounter{appendixctr}
\renewcommand{\theappendixctr}{\Alph{appendixctr}}
\newcommand{\appendixsection}[1]{
  \refstepcounter{appendixctr}
  \section*{Appendix \theappendixctr: #1}
}

\def\month{07}  
\def\year{2026} 
\def\openreview{\url{https://openreview.net/forum?id=wkCTCXo1a9}}

\usepackage{hyperref}  
\usepackage{url}            
\usepackage{booktabs}       
\usepackage{amsmath}        
\usepackage{amsfonts}       
\usepackage{nicefrac}       
\usepackage{microtype}      
\usepackage{xcolor}         
\usepackage{graphicx} 
\usepackage{changepage}  
\usepackage{rotating}
\usepackage{algorithm}
\usepackage{algorithmic}
\usepackage{listings}
\usepackage{arydshln}
\usepackage{multirow}

\title{Progressive Checkerboards for Autoregressive \\ Multiscale Image Generation}

\author{David Eigen \email de@deigen.net}

\begin{document}

\maketitle

\begin{abstract}
A key challenge in autoregressive image generation is to efficiently sample
independent locations in parallel, while still modeling mutual dependencies
with serial conditioning.  Some recent works have addressed this by
conditioning between scales in a multiscale pyramid.  Others have looked at
parallelizing samples in a single image using regular partitions or randomized
orders.  In this work we examine a flexible, fixed ordering based on
\emph{progressive checkerboards} for multiscale autoregressive image
generation.  Our ordering draws samples in parallel from evenly spaced regions
at each scale, maintaining full balance in all levels of a quadtree subdivision
at each step.  This enables effective conditioning both between and within scales.
Intriguingly, we find evidence that in our balanced setting, a
wide range of scale-up factors lead to similar results, so long as the total
number of serial steps is constant.  On class-conditional ImageNet, our method
achieves competitive performance compared to recent state-of-the-art autoregressive
systems with like model capacity, using fewer sampling steps.
\ifanonymous
Code will be made available.
\else
Code is available at \url{https://github.com/deigen/checkerboardgen}.
\fi
\end{abstract}

\begin{figure}[h]
\vspace{-1mm}
	\centering
	\includegraphics[width=0.95\textwidth]{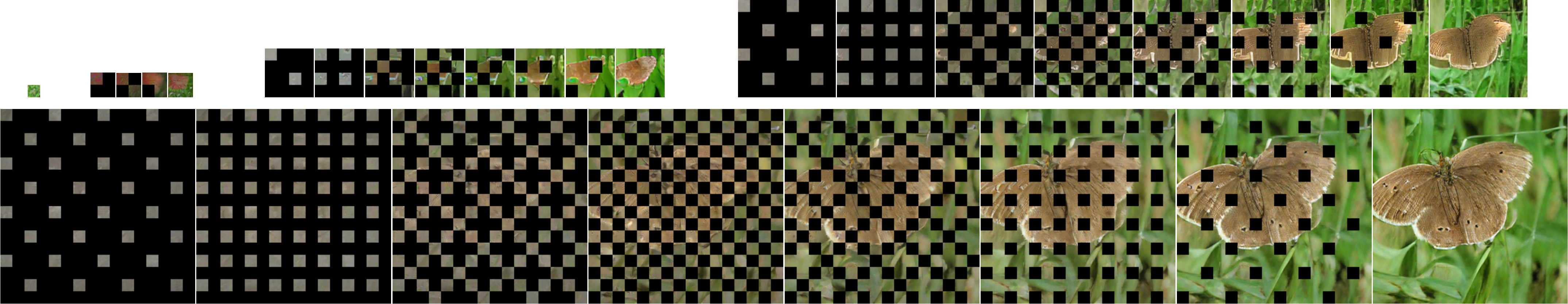} \\
	\caption{
	Progressive checkerboard samples from our model using 2x scale factor and 8 steps per scale.
	Masking applied to sampled locations at each step after decoding for visualization.
	}
	\label{fig:teaser}
\vspace{-2mm}
\end{figure}

\section{Introduction}
\vspace{-1mm}

Image generation methods, including autoregressive (AR) models, diffusion and
masked models, have shown impressive results in generating high-quality images
\citep{dalle,dalle2,imagen,pixelcnn,pixelcnn++,vqvae2,ho2020diffusion,latent-diffusion,pmlr-v202-chang23b,yu2024language}.
AR models generate images by sampling a sequence of image tokens from
a learned distribution, where each token is conditioned on previously sampled tokens.
Recent AR methods include scalewise autoregressive models
\citep{var,infinity,star}, which condition from coarse to fine as their
progressive sequence, as well as parallel AR models
\citep{par2025,zhang_locality-aware_2025}, which select multiple locations at once
by conditioning with sequential blocks.  Both conditioning methods aim to model
dependencies between tokens in a way that allows for sampling
multiple locations in parallel, which is critical for fast generation.

However, samples drawn in parallel
are also drawn independently from one another, which can
produce inconsistencies when they are mutually dependent.  This is particularly true for
adjacent and nearby locations.  For example, if an image patch is red, the one
next to it is likely to be red as well, but drawing tokens independently may choose
two different colors instead of the same one.
Conditioning each location on a common parent, for example in the previous scale of a scalewise progression, makes the conditional
samples more independent of each other \citep{var}.

Relying exclusively on scalewise conditioning, however, depends on a
slow scale-up factor:  If the scale factor is too large, then an object
spanning multiple locations in one scale may not yet be visible in the previous scale,
so its appearance dependencies would not be conditioned and
independently sampling can mix modes.
Rather than scaling up by a factor of 2, 
the best performing multiscale models \citep{var,infinity,star} currently scale by a factor of
$\sqrt[3]{2} \approx 1.26$ (i.e. each factor of 2 is further subdivided into three scales).
However, as prior sequential and parallel methods \citep{par2025,zhang_locality-aware_2025}
have shown, another way to condition is to model dependencies between sampling locations.

In this paper, we examine conditioning between locations \emph{within} each
scale, making use of both between- and within-scale conditioning.  We develop a
sampling order based on a progressive checkerboard, which reduces mutual
dependence within each sampling block and enables fast scale-up.  Our
ordering maintains balance at all levels in a quadtree subdivision, so that
varying the block size effectively varies the trade-off between parallelism and
modeling conditional dependencies.  We use this to explore the relationships
between scale-wise and within-scale conditioning, finding somewhat surprisingly
that for this spatially balanced setup, 
the total number of sequential steps largely determines performance
independently of how the steps are divided among scales.
Our model is competitive with state-of-the-art methods in ImageNet 256x256
class-conditioned generation, using just 17 sampling steps.

\section{Background and Related Work}
\vspace{-1mm}
\subsection{Background:  Autoregression for Images}
\vspace{-2mm}

Autoregressive (AR) models for
images sample a sequence of tokens from a learned distribution, where each
token is conditioned on previously sampled ones.  Tokens representing images
are typically formed using a VAE autoencoder
\citep{vqvae2,latent-diffusion}, which maps between RGB images and
local patch representations that are clustered into a discrete set of codes.
The AR model then learns distributions over token codes $z_t$.  In the simplest case, this
is done one at a time in a conditional sequence,
$
P(z_1, z_2, \ldots, z_N) = P(z_1) P(z_2 | z_1) \ldots P(z_N | z_1, z_2, \ldots, z_{N-1}).
$
Typically, the distribution is modeled using a transformer \citep{vaswani2017attention,gpt,gpt2}.
To generate an
image, we draw a sequence of $N$ tokens from the model, where each token $t$ is sampled 
from $P(z_t|z_1,\ldots, z_{t-1})$.
The VAE decoder converts the token representations into an RGB image.
However, sampling tokens one at a time can be slow.  Thus, recent methods have explored
ways to speed up sampling by computing multiple tokens in parallel.

\vspace{-1mm}
\subsection{Scalewise Autoregression}
\vspace{-2mm}

Gradual scaling AR models \citep{var,star,infinity},
introduced by VAR \citep{var}, progressively develop a
full-resolution array of latent codes, which is used as a working ``canvas'' to
apply residuals.  At each iteration, the canvas is
downsampled to the current scale (which grows by a factor of $\sqrt[3]{2}$)
and a transformer predicts residuals, which
are upsampled and applied to the full-resolution array.

The downsample-predict-upsample cycle implicitly invokes a progressive
deblurring in latent space, similar to the progressive denoising of diffusion models.
Viewing the method this way sheds some light on reasons why the scaling factor
remains small.  Just as diffusion models struggle with large denoising steps
due to their use of independent sampling and Gaussian noise
 \citep{ddim,progressive-distillation,consistency-models,dpm-solver},
scalewise AR models can struggle with
large deblurring steps due to independently sampling underlying multi-modal
distributions \citep{cold-diffusion,blurring-diffusion}.

\vspace{-1mm}
\subsection{Parallel Sampling}
\vspace{-2mm}

Parallel autoregression (PAR) by \citet{par2025} generates
images by sampling in parallel from equal-sized square partitions.  Their
method uses raster order within each partition and no multi-scale sampling,
limiting its parallelism to only four groups.
More recently, \citet{zhang_locality-aware_2025} use a
locality-aware ordering that adds far-away locations while 
growing already-selected regions with adjacent samples, while
\citet{randar2025} use random orderings.  However, the former uses a more
complex dynamic evolution to grow regions, while both methods require
additional position probe tokens, extending the overall sequence length.  We
use a simple but effective regular pattern with multiscale
conditioning, and do not require probe tokens.

\citet{yan_gtr_2025} use a two-stage ``generation then
reconstruction'' approach applied in the context of MAR \citep{mar2024}
that applies unmasking using diagonal striations, the final stage of which corresponds to
a mod 2 checkerboard.  In contrast to their method, we use a fully balanced
progressive checkerboard, along with explicit
multiscale conditioning, in the context of direct autoregression instead of masking.
Additional recent works include 
xAR \citep{xar2025}, which uses flow-matching 
to generate large chunks at each AR step;
NAR \citep{nar2025}, which uses horizontal and vertical decoders to
condition in striations; and ARPG
\citep{arpg2025}, which uses cross-attention to condition random-order
samples with masked models.

Some gradual scaling models also incorporate ways to reduce same-scale
dependencies.  \citet{infinity} use random
bit flips to address errors from independent
sampling at the quantizer bit level after incorporating BSQ
\citep{zhao2024bsqvit} as their tokenizer.  \citet{star} use a
miniature masked image model applied in the autoregressor head to condition
lower-confidence locations on higher-confidence ones.  To
enable 2x scaling, \citet{flowar2025} stack a flow-matching
model over mixed-mode AR samples to resolve interdependences.
\citet{kutscher_reordering_2025} examine the impact of patch
ordering in the context of recognition models.

\section{Method}
\vspace{-0.5em}

\subsection{Autoregressor}
\label{sec:ar-arch}

\vspace{-10mm}
\begin{figure}[h]
	\centering
	\includegraphics[width=0.42\textwidth]{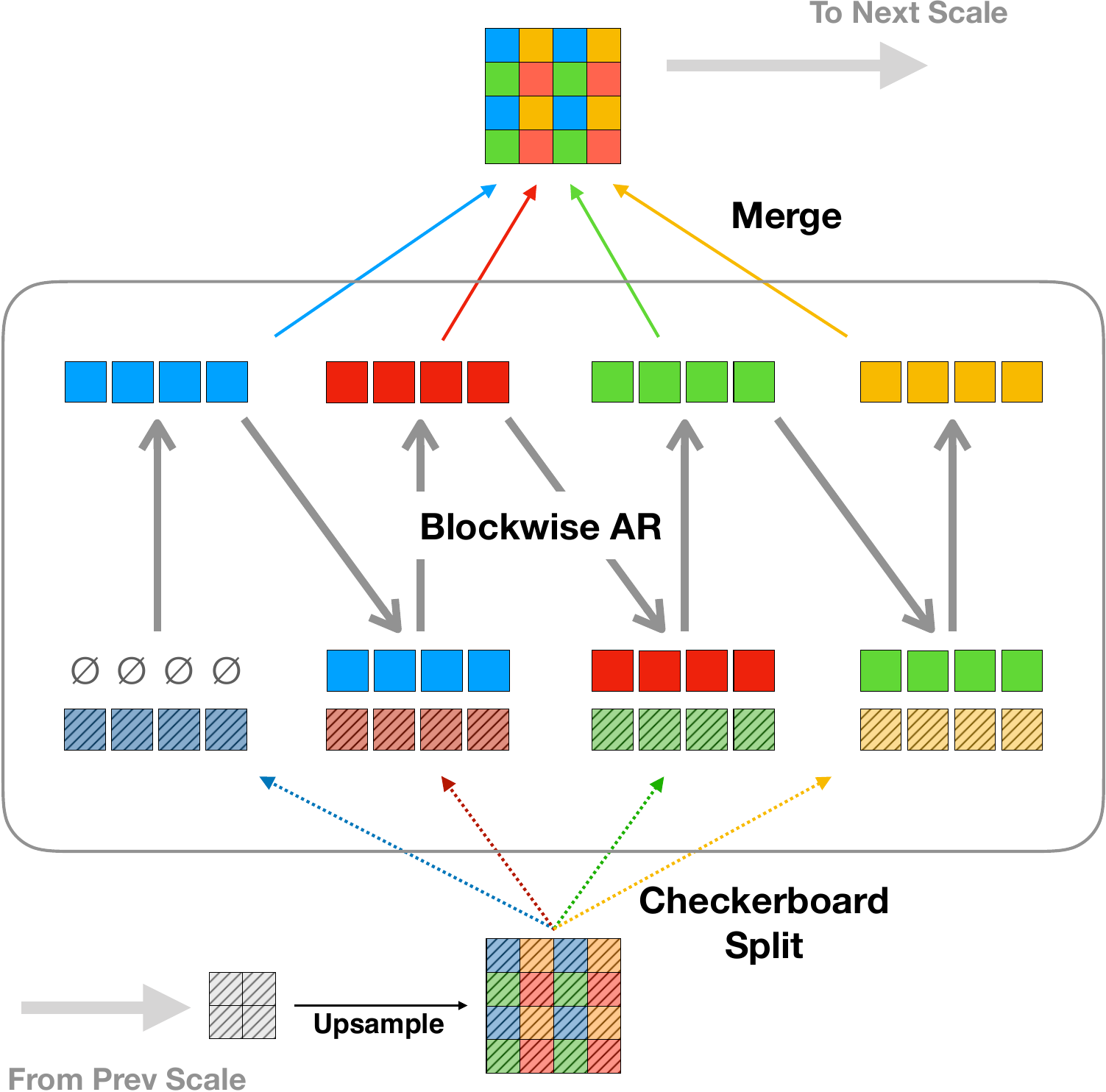}
	
	\caption{
	Overview of our multiscale blockwise checkerboard autoregressor.
	}
	\label{fig:ar-arch}
\end{figure}

Our multiscale progressive checkerboard autoregressive model is built around a transformer
\citep{vaswani2017attention,gpt,gpt2}
with blockwise causal mask using successive checkerboard sampling blocks.

Figure \ref{fig:ar-arch} provides an overview of our method.  At each scale
$s$, latent codes output by the previous scale $s-1$ are upsampled by a scaling factor $r$ to form the
scalewise-conditioning input, $z^{up}_{s-1} = {\rm upsample}(z_{s-1}, r)$.  The locations
of this map are split according to the progressive checkerboard ordering
(see Sec. \ref{sec:progressive-checkerboard}) into $P$ blocks, $b_s^1, b_s^2, \ldots, b_s^P$,
where each block $b_s^i$ contains $H_s W_s / P$ tokens.
We randomize $P$ during training to allow different degrees of parallelism
at inference time.  

In order for each autoregressive step to condition on previously sampled tokens
\emph{within} each scale, we include the output of each block into the input of
later ones.  In particular, we use a linear combination between the
upsampled values $z^{up}_{s-1}$ from the previous scale and the (shifted)
outputs $z_{s}$ of the current scale, along with learned position embeddings $pos$ at
both locations.  The input tokens to the transformer are then
\begin{equation}
inputs(s, i) = W_{proj} \cdot {\rm Concat}(z^{up}_{s-1}[b_s^i], z_s[b_s^{i-1}], pos[b_s^i], pos[b_s^{i-1}])
\label{eq:inputs}
\end{equation}
where $W_{proj}$ is a learned linear projection, and $[\cdot]$ indicates
indexing at the specified block locations.  Since the first block $b_s^1$ has no
previous block, we use learned constant vectors for $z_s[b_s^0]$ and $pos[b_s^0]$.

At inference, each block $b_s^i$ contains the set of locations for each sampling step.
Tokens are processed in parallel within each block, while the blocks
themselves are serialized and sampled sequentially.
Output tokens for each block are sampled independently using a multinomial.

During training, the entire sequence is processed in parallel using the ground
truth codes $z_s$ for both inputs and targets autoregressively.
We use a blockwise causal mask where tokens within each block
$b_s^i$ can attend to each other, as well as to tokens from all previous blocks,
both for the same scale and in lower-resolution scales.  Each token position
outputs softmax logits for a quantized codebook; we use flat mean cross-entropy loss
over all scales and positions.

Importantly, the locations in each block are spaced as evenly as
possible.  The number of locations that can be sampled in parallel
depends on the independence between locations, when conditioned both on the
previous scale and samples in the current scale.  Thus, there is a trade-off 
between the number of scales and the degree of parallelism within each scale, a
relation that we explore later in Sec \ref{sec:scale-steps-evals}.

\subsection{Progressive Scan Order}
\label{sec:progressive-checkerboard}

\begin{figure}[h]
	\centering
	\includegraphics[width=0.95\textwidth]{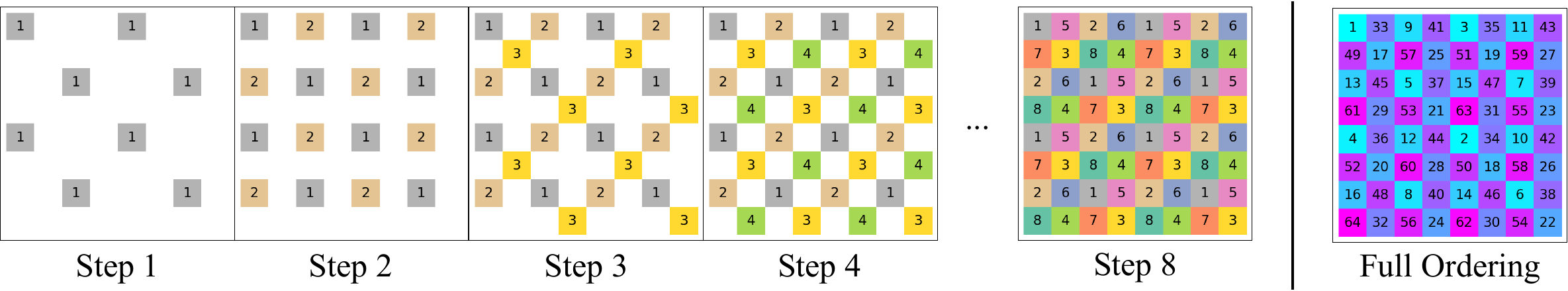} \\
	\vspace{-0.5em}
	\caption{
	(a) Progressive checkerboard on an $8\times8$ grid using $P=8$ steps.
	(b) Full ordering.
	}
	\label{fig:progressive-checkerboard}
\end{figure}

\lstdefinestyle{mystyle}{
	basicstyle=\ttfamily,
	keywordstyle=\color{blue}\ttfamily,
	numberstyle=\ttfamily\small,
	columns=fullflexible,
	keepspaces=true,
	xleftmargin=-1em,
	tabsize=4
}
\lstset{style=mystyle}

\vspace{-0.25em}
\begin{algorithm}[h]
\caption{Progressive Checkerboard Scan Order}
\vspace{-0.25em}
\label{alg:scanorder}
\begin{lstlisting}[language=Python]
  def ProgressiveCheckerboard(size, x=0, y=0):
      if size == 1: return [(x, y)]
      d = size // 2
      # create balanced lists for TL, BR, TR, BL
      sublists = [
          ProgressiveCheckerboard(d, xi, yi)
          for (xi, yi) in ((x,y), (x+d,y+d), (x+d,y), (x,y+d))
      ]
      # combine round-robin from the quadrants
      return concat(zip(*sublists))
\end{lstlisting}
\end{algorithm}

To create the progressive checkerboard ordering, we use a divide-and-conquer
approach, where the 2D grid of size $N\times N$ is recursively subdivided
into quadrants, and locations selected to maintain a spatially balanced
assignment at all quadtree levels.
At each recursion step, balanced index lists are generated for each sub-quadrant
recursively, then merged using round-robin selection with a diagonal skip-step pattern
(i.e., TL, BR, TR, BL).  This results in a spatially balanced progressive order,
with a unique index in $1\dots N^2$ assigned to each of the $N\times N$ spatial
positions.  (If the grid size $N$ is not a power of 2, we generate the order on the next power of 2
and restrict to the grid size).  See Algorithm \ref{alg:scanorder}.

This scan order can be used flexibly with various
partition sizes.  To create the blocks $b_s^i$ for each scale $s$, we simply
divide the progressive scan order into $P$ contiguous segments of equal
size.  Figure
\ref{fig:progressive-checkerboard} shows an example progressive checkerboard
ordering on an $8\times 8$ grid.

\subsection{Token Embeddings}
\label{sec:embeddings}

For token embeddings, we use quantized latent codes from a VAE-based
autoencoder \citep{autoencoding-better,taming-transformers,dalle,latent-diffusion}.
Each scale is encoded separately from the RGB image, so that latents
are directly interpretable and decodable.
For transformer inputs, we pass the VAE latent codes through a small MLP to
unfold them from their low-dimensional clustering space back to higher
dimension.
In addition, we use a
class token as the first input, which along with AdaLN \citep{dit23} is used to
condition on class labels with classifier-free guidance
\citep{cfg22}.

\subsection{Position Encodings and RoPE Mixing}

To make use of our blockwise checkerboard order's ability to model
local dependencies,
the transformer must be able to attend to adjacent locations for each token.
We accomplish this using both input position embeddings and learned
rotary encodings (RoPE) \citep{rope,rope2d}.
Embeddings encode the position in the image-scale space, not the sampling
order.
We initialize RoPE using factored space and scale
representations (7/8 spatial, 1/8 scale).
The embeddings are then learned in the joint space,
allowing the model to find a high-frequency adjacent-position basis and other
attention patterns optimized for our sampling order.

Each token in the transformer
processes inputs from two locations:
the current sampling positions $b_s^i$, and the conditioning latents at the
previous block's positions $b_s^{i-1}$ (Sec. \ref{sec:ar-arch}).
Eq. (\ref{eq:inputs}) supplies both in the input layer mixture.
However, without modification, the attention layers only use RoPEs for the
current positions $b_s^i$.
To enable attention on both sets of positions, we looked at mixing RoPEs
for attention keys using learned 
mixing coefficients\footnote{We learn the ``logits'' $\alpha'$ of the mixing coefficients
and set $\alpha_{lh} = {\rm sigmoid}(\alpha'_{lh})$.}
$\alpha_{lh}$ for each transformer layer $l$ and attention head $h$:
$$
rope_{lh}(b_s^i) = \alpha_{lh} \cdot rope(b_s^i) +
(1 - \alpha_{lh}) \cdot rope(b_s^{i-1}) ,
$$
where $rope(\cdot)$ produces RoPE embeddings for specified locations. Mixing is
only applied to keys; queries always use the positions $b_s^i$,
corresponding to the current sampling locations whose output is being computed.
Although we did not see any performance gains with this
strategy, we examine its behavior in Sec. \ref{sec:rope-mixing-exper}, finding that
only the first two layers use the previous block's RoPE shifts. This indicates that conditional
information is extracted early on, so input mixing is sufficient.

\section{Experiments}

\subsection{Model and Training Details}

We train our models on ImageNet \citep{imagenet} at 256x256 resolution, using two model
sizes:  the small (S) model has 12 transformer layers with hidden dimension 512 and 16
attention heads, while the large (L) model has 20 layers with 1024 hiddens and 16 heads.
In most ablation experiments we use the S model, trained for 100 epochs with batch
size 128 and learning rate $1\times10^{-4}$, dropped to $1\times10^{-5}$ in two steps
over 10 epochs.  The L model is trained for 200 epochs with batch size 64 and
learning rate $5\times10^{-5}$, dropped to $1\times10^{-5}$ for 5 epochs, followed by
a second effective drop by increasing batch size to 320 via gradient accumulation for the
last 5 epochs.  In all cases, we use AdamW \citep{adamw} optimizer with weight decay
0.01 and ten-crop transforms, running on a single NVIDIA GH200 GPU.

We use the VAE autoencoder from LlamaGen \citep{llamagen_2024}.  To better
represent smaller image sizes in our multiscale model, we fine-tune by
freezing all layers other than the quantizer codebook itself, and retrain the
just the codebook layer with a size of 4096 on ImageNet for one epoch, using
random image sizes between 16 and 256 and L2 loss on the latent codes.
To measure sample quality, we compute Frechet Inception Distance (FID) \citep{fid}
and Inception Score (IS) \citep{inception-score}, using 50k samples
and standard reference set using the TensorFlow implementation from
\citet{llamagen_2024,dhariwal2021diffusion}.

For classifier-free guidance (CFG), we found applying CFG too early in the
sampling progression limits diversity, likely due to the first samples'
influence on global structure.  Because of this, we sample the first 5 steps
(corresponding to the 1x1 and 2x2 scales when scaling by 2x) with CFG=0, and apply CFG for all subsequent steps.
Following common practice \citep{par2025,nar2025,zhang_locality-aware_2025}, we
perform a sweep of CFG values at 0.1 increments using the reference set.

\subsection{Benchmark Comparison}

\newcommand{\tabbf}[1]{{\bf #1}}
\newcommand{\tabit}[1]{\underline{#1}}
\begin{table}[t]
\vspace{-2ex}
\centering
\small
\begin{tabular}{lcc|cccc|cc}
\hline
Model & Type/Tok & Params & FID $\downarrow$ & IS $\uparrow$ & Pre.$\uparrow$ & Rec.$\uparrow$ & \#Steps & Time (s) \\
\hline
\hline
DiT-XL/2 \citep{dit23} & Diffu-KL & 675M & 2.24 & 278.2 & 0.83 & 0.57 &  ~$1\times250$ & 11.9 \\
MAR-L \citep{mar2024} & MAR-KL & 479M & 1.78 & 296.0 & 0.81 & 0.60 & $64\times100$ & 26.4 \\
GtR \citep{yan_gtr_2025} & MAR-KL & 479M & 1.81 & 297.4 & --- & --- & $32\times30$ & --- \\
xAR \citep{xar2025} & Flow-KL & 608M & 1.28 & 292.5 & 0.82 & 0.62 & $4\times50$ & 7.7 \\
\hline \hline
LlamaGen-L \citep{llamagen_2024} & AR-VQ & 343M & 3.07 & 256.1 & 0.83 & 0.52 & 576 & 12.58 \\
VAR-d16 \citep{var} & VAR-VQ & 310M & 3.30 & 274.4 & 0.84 & 0.51 & \tabbf{10} & \tabbf{0.12} \\
PAR-L-4x \citep{par2025} & AR-VQ  & 343M & 3.76 & 218.9 & 0.84 & 0.50 & 147 & 3.38 \\
RandAR-L \citep{randar2025} & AR-VQ  & 343M & 2.55 & 288.8 & 0.81 & 0.58 & 88 & 1.97 \\
NAR-L \citep{nar2025} & AR-VQ & 372M & 3.06 & 263.9 & 0.81 & 0.53 &  31 & 1.01 \\
ARPG-L \citep{arpg2025} & AR-VQ  & 320M & \tabbf{2.30} & \tabit{297.7} & 0.82 & 0.56 & 32 & 0.58 \\
LPD-L \citep{zhang_locality-aware_2025} & AR-VQ & 337M & \tabit{2.40} & 284.5 & 0.81 & 0.57 &  20 & \tabit{0.28} \\[0.2mm]
\hline
\hline
Checkerboard-L 2x cfg=1.4 (Ours) & AR-VQ & 343M & 2.72 & \tabbf{302.5} & 0.81 & 0.56 & \tabit{17} & 0.52 \\
Checkerboard-L 2x cfg=1.5 (Ours) & AR-VQ & 343M & 2.83 & \tabbf{318.2} & 0.82 & 0.57 & \tabit{17} & 0.52 \\
\hline
Checkerboard-L 4x cfg=1.7 (Ours) & AR-VQ & 343M & 2.79 & \tabbf{311.5} & 0.80 & 0.57 & \tabit{17} & 0.52 \\
\hline
\end{tabular}
\caption{Image generation models on ImageNet 256x256.  We mainly compare to other AR-VQ
methods (bottom rows) that are directly comparable, but also include non-VQ methods for additional context (top, $\times$ indicates mask/ar and diffusion/flow steps).
Inference time measured for single image generation on A100.\\
}
\label{tab:benchmarks}
\end{table}
\begin{figure}[t]
\vspace{-3mm}
\centering
\raisebox{3ex}{(a)}
\raisebox{-5ex}{
\includegraphics[width=0.2\textwidth]{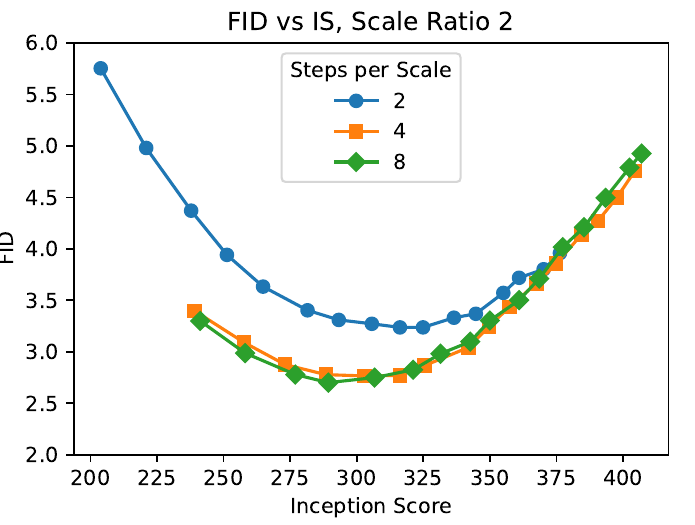}
}
\raisebox{3ex}{
(b)
{\small
\begin{tabular}{l|l}
Ratio & Sizes (\#patches/side) \\
$\sqrt{2}$  & [1, 2, 3, 4, 6, 8, 11, 16] \\
2 & [1, 2, 4, 8, 16] \\
3 & [1, 2, 5, 16] \\
4 &  [1, 4, 16]
\end{tabular}
}
(c)
{\small
\begin{tabular}{lcc}
  \hline
{\bf Sampling Order}& {\bf FID} \\
  \hline
Random S 2x & 6.01 \\ 
Checkerboard S 2x & 5.32  \\
  \hline
\end{tabular}
}
}
\caption{
(a) FID vs IS computed at 0.1 CFG increments. L model size at 2, 4 and 8 steps per scale. \\
(b) Exact scale sizes used in our experiments, for 256x256 image and 16x16 VAE patch size.
(c) Checkerboard vs random order, S model at 2x scale ratio.  Checkerboard performs better due to its spatially balanced order.
}
\label{fig:fid-iscore-L}
\end{figure}

Table \ref{tab:benchmarks} compares our method against recent autoregression-based image generation
models of similar model size on ImageNet 256x256.  We show results for our Checkerboard-L
model with a 2x scaling ratio and 4 sampling steps per scale (17 total), as well as a 4x ratio with 8
steps per scale (also 17 total)\footnote{For the L models, the 2x model is trained from scratch for 200 epochs, and the
4x model is initialized with the 2x model weights and trained for 30 epochs.}.
Directly relevant comparisons are with other autoregression-based methods with VQ sampling,
shown with ``AR-VQ'' in the second column.  We also show methods that use non-quantized
autoencoders for wider context, though these are not directly comparable.
Qualitative samples are in Fig. \ref{fig:samples}, and selected failures
in Appendix \ref{app:failurecases}.

Compared to PAR \citep{par2025} and
RandAR \citep{randar2025}, to which our method is most closely related, we achieve
\emph{similar or better FID and IS with fewer sampling steps and faster inference time}.  While
PAR uses 147 steps and RandAR 88 steps, we only need 17 steps to achieve a FID of 2.72
(compared to 3.76 and 2.55, respectively).  Our inference time of 0.52s per image is also
faster than both methods (3.38s for PAR and 1.97s for RandAR) on A100.

In addition, our method is competitive with the more recent AR-based models
ARPG \citep{arpg2025} and LPD \citep{zhang_locality-aware_2025}, notably also
using fewer steps, though FID is somewhat lower.  Although the reasons for
lower FID are unclear, two possibilities are: (\emph{i}) Training uses batch
size 64 for 200 epochs (increased to 320 for the last 5) while LPD and ARPG
use 2048 and 1024, respectively, for 400 epochs.  Thus, our
schedule has fewer total examples and noisier gradients.  And (\emph{ii})
ARPG and LPD separate tokens for sampled values and queries, while we combine them in an
input mixture.  This results in half as many tokens during training, but at
the cost of using some KV cache capacity for query information that may not be useful
for later steps.  Combining strengths of all three approaches is a promising
direction for future work.

We also verify the effectiveness of our checkerboard ordering in Figure
\ref{fig:fid-iscore-L}(c): using the S model at 2x scale ratio, we train a
comparison model using random within-scale order.  The checkerboard performs 
substantially better, due to its spatially balanced sampling pattern.
In addition, we show the full CFG sweep for our L model in Fig.
\ref{fig:fid-iscore-L}(a).
While only 2 steps per scale is 
ineffective, both 4 and 8 perform similarly.
In the next section, we explore the impact
of number of steps in greater detail, and how these relate to the scaling
ratio.

\subsection{Relationship Between Scale Ratio and Sampling Steps}
\label{sec:scale-steps-evals}

To evaluate and compare different scale factors
and sampling steps, we train our small (S) sized model using
four scale-up factors: $\sqrt{2}, 2, 3$
and $4$, as well as a single-scale baseline
(exact sizes in Fig \ref{fig:fid-iscore-L}(b)).
We then evaluate each one using different numbers of
sampling steps, varying the number of steps per scale, which corresponds to
partitioning using the checkerboard pattern.  Note that in our implementation,
each scale ratio requires retraining, but the number of sampling steps can
be varied at inference.

\begin{figure}[t]
	\centering
	\vspace{-2ex}
	\includegraphics[width=0.45\textwidth]{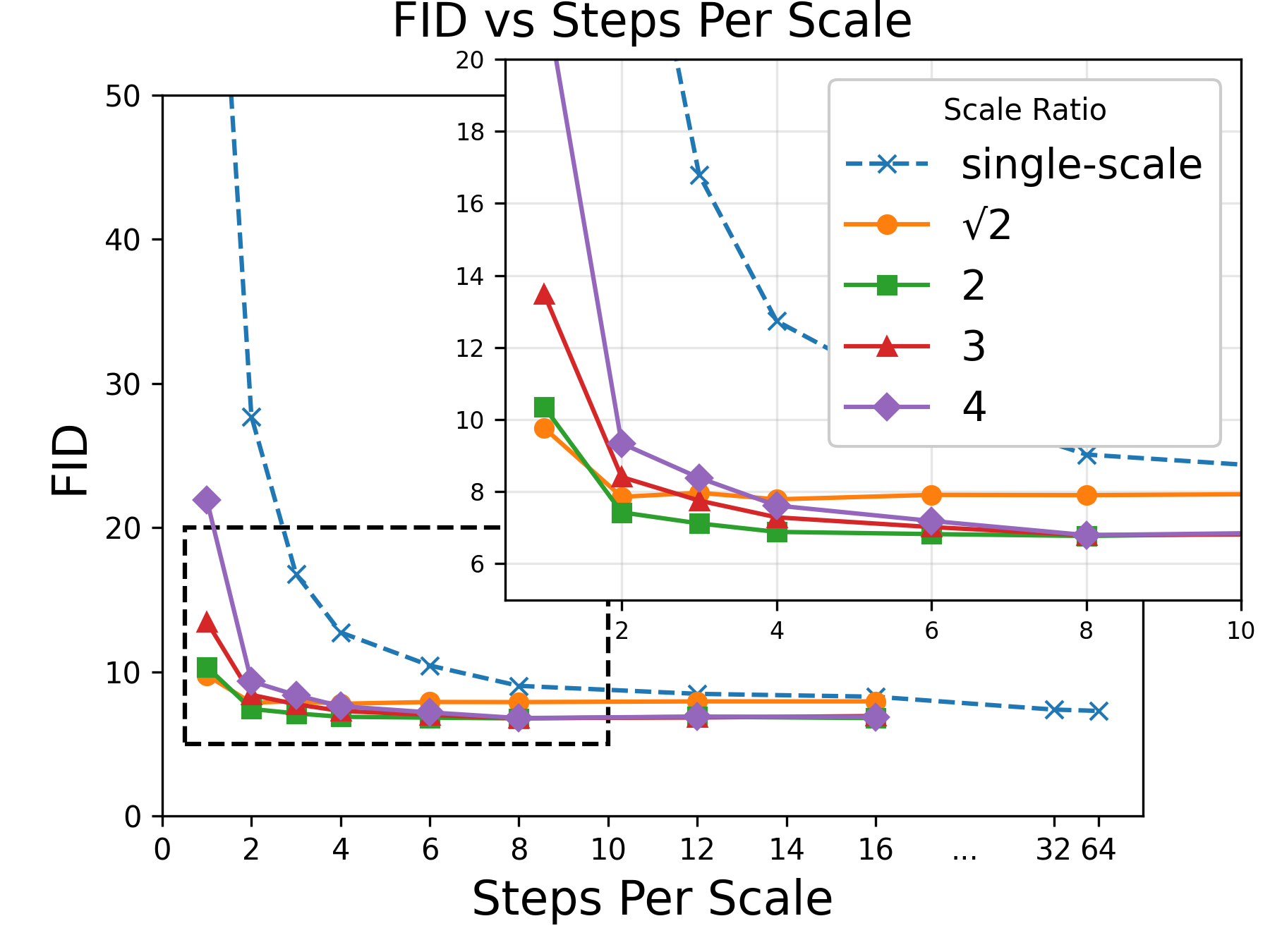}
	\includegraphics[width=0.45\textwidth]{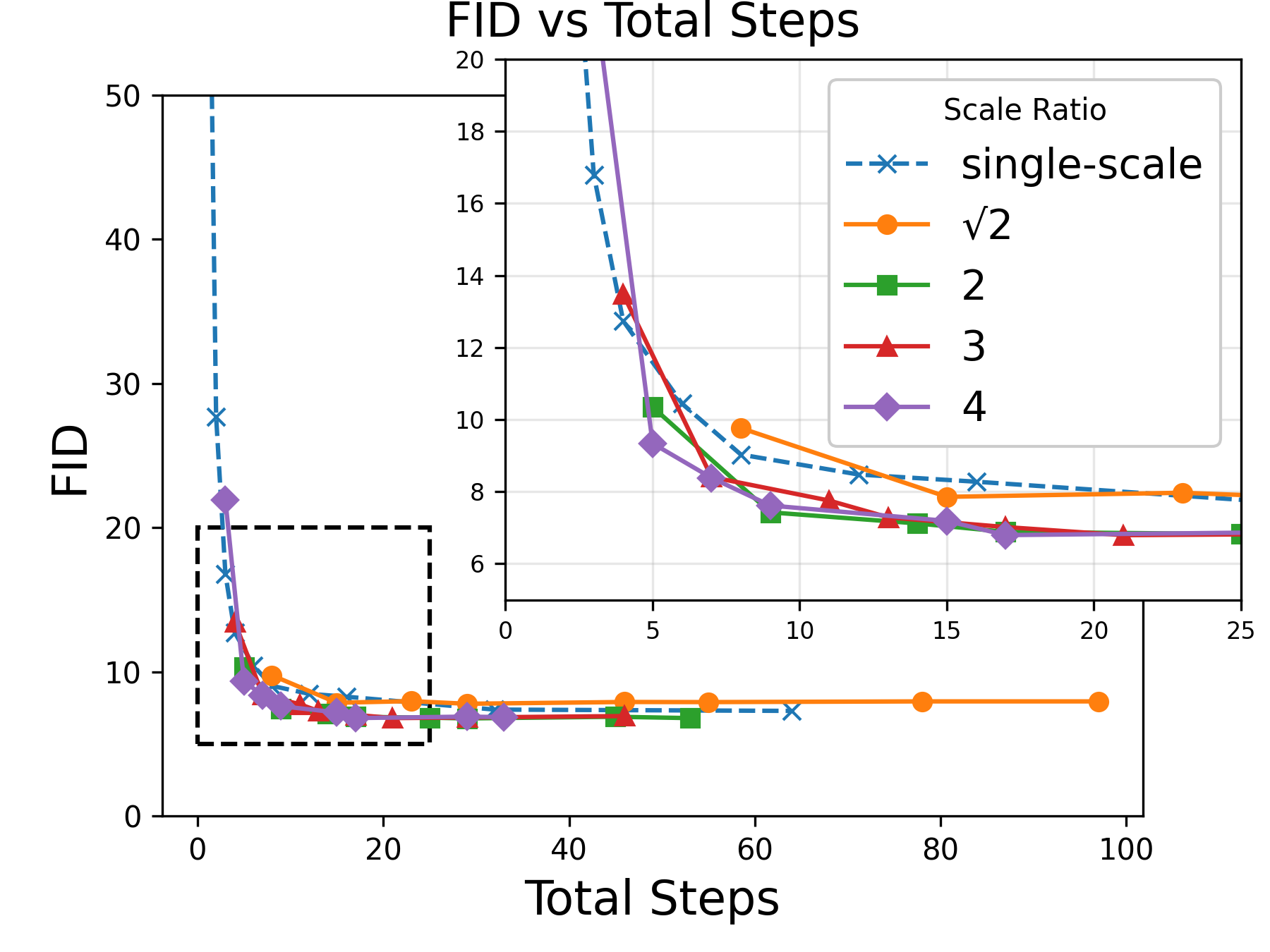}
	\\
	\includegraphics[width=0.44\textwidth]{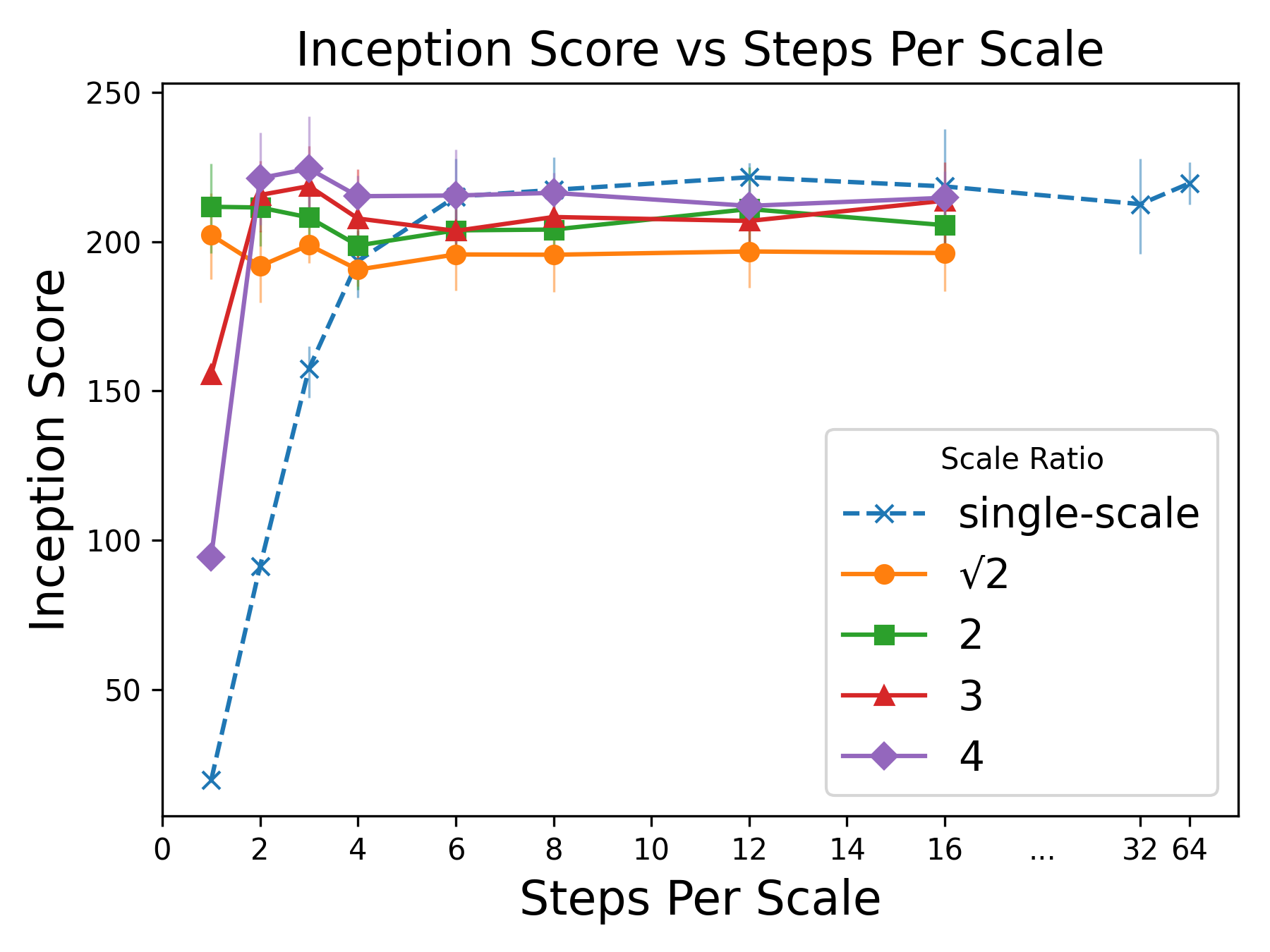}
	\hspace{0.5mm}
	\includegraphics[width=0.44\textwidth]{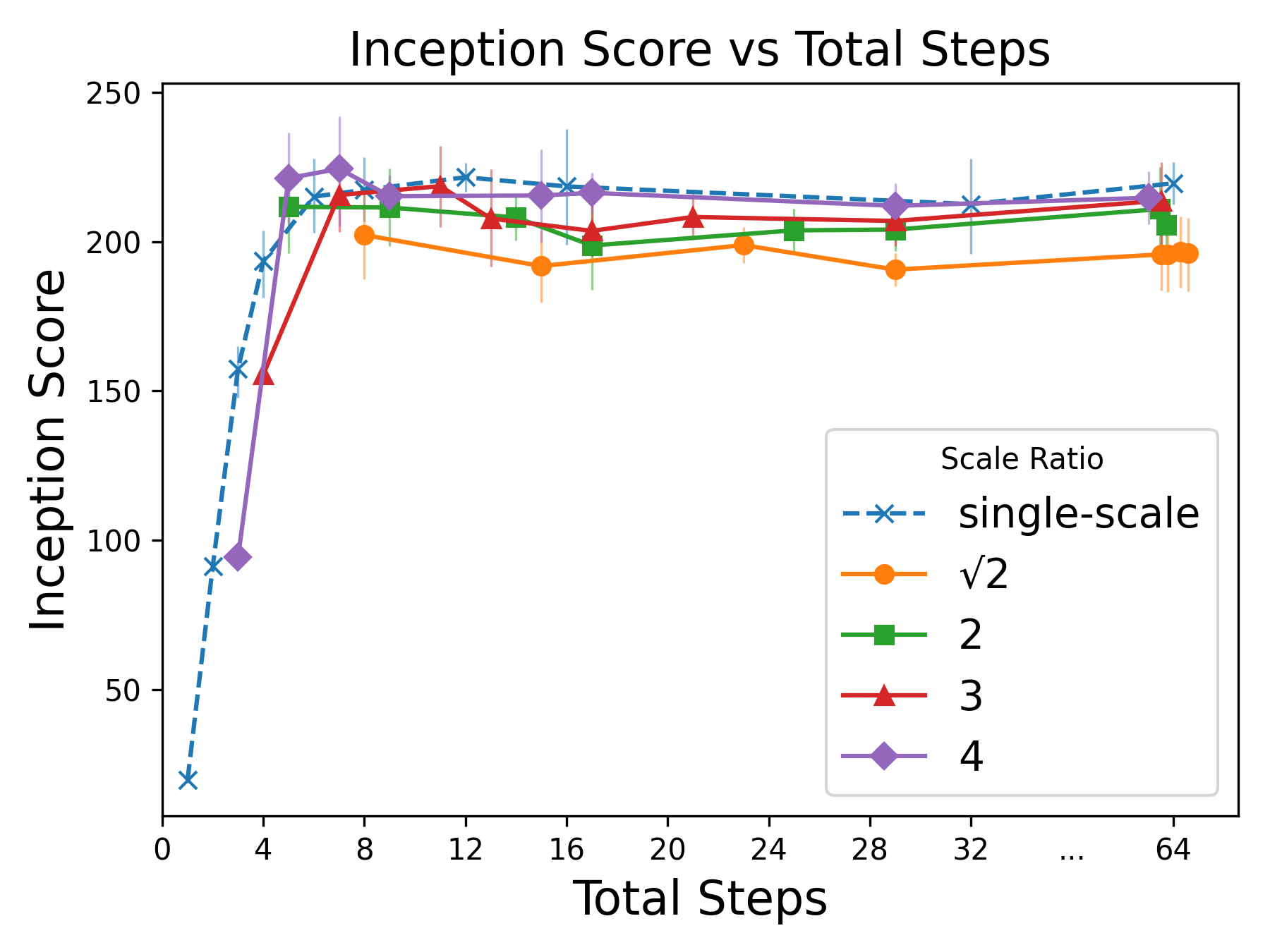}
	\hspace{0.0mm}
	\vspace{-1mm}
	\caption{
	FID (top) and IS (bottom), by scale ratio and number of inference steps for the S model size.
	Left: by numbers of steps per scale; Right: by total number of steps.
	Multiscale models outperform the single-scale baseline.
	Note that scale ratios 2, 3 and 4 all achieve similar performance counted by \emph{total} steps,
	even though the steps per scale needed to reach each total is different for each scale factor.
	}
	\vspace{-3mm}
	\label{fig:scale-ratios}
\end{figure}

We evaluate each condition using FID and IS with 50,000 samples, sweeping
classifier-free guidance in 0.1 increments to obtain the best FID value for
each setting.  CFG is applied at all scales.  To account for
small variations in evaluation and sampling, for each point we record 
values with FID within 2\% (relative) of its minimum, and plot the mean, min
and max IS of these with error bars.

Figure \ref{fig:scale-ratios} shows the results.  On the left, we plot FID and IS
by number of steps per scale, while on the right we plot them by total number of steps.
While IS scores are similar between multiscale models and the single-scale baseline,
all multiscale models outperform the single-scale baseline in terms of FID, with the best
performance achieved at scale ratios 2, 3 and 4.
As expected, larger scale-up factors require more steps per scale to achieve good performance
(left-side figures).

However, when viewed by the \emph{total} number of steps, scale ratios 2, 3 and 4 all
achieve similar performance, \emph{even though each scale ratio
uses a different number of steps per scale} to reach the total.  The lines for these three
ratios all overlap tightly (Fig. \ref{fig:scale-ratios}, top-right).  In addition,
all outperform the single-scale baseline as well as slower $\sqrt{2}$ scaling.
This indicates that while multiscale conditioning is important, the exact scale
factor is not very sensitive:
there are multiple ways to allocate a given number of steps among scales
and obtain similar performance.
Indeed, our checkerboard order also includes a
degree of coarse-to-fine conditioning, simply due to its spatially balanced
sampling, which may contribute to this effect (see Appendix \ref{app:scale-steps-mechanism}).
Thus, for our balanced multiscale generation, \emph{the total number of steps in the conditional chain is the dominant
performance factor}.

Moreover, best performance is achieved at around 17 total steps for all of the
multiscale models, while steps beyond this add little if any benefit.  This
makes sense, since partitioning our balanced checkerboard order into more steps
spaces out the locations within each step, and coarse-to-fine conditioning is
also included in the multiscale pyramid.  Within-scale steps are needed
largely to condition within the local upsampling windows.
However, while
between- and within-scale conditioning can model overlapping dependencies, we
still find benefit to including both.

Additionally, we verify our findings at 512 resolution in
Appendix~\ref{app:expers512}, and for the L model at 256 in
Appendix~\ref{app:scale-steps-L256}.

\subsection{RoPE Mixing}
\vspace{-1mm}
\label{sec:rope-mixing-exper}

As described in Sec. \ref{sec:embeddings}, we experimented with mixing RoPE
embeddings for attention keys using learned mixing coefficients $\alpha_{lh}$
for each layer $l$ and head $h$.  Table \ref{tab:rope-mixing}(a) compares
mixing for all layers, no layers, and only the first two layers.
Training time
increases when enabled for all layers, but is negligible
when applied just to the first two.
While we found no significant differences in FID or IS, the behavior of the
mixing coefficients shown in \ref{tab:rope-mixing}(b) is illustrative of how
the model uses sampling information in the inputs.  Only the
first two layers attend to
sampled positions, indicated by negative weights.  This suggests the model may extract 
information from the sampled locations early, possibly
embedding it into representations at relevant output positions, and relying on
attention based on output positions for most of the transformer.
While this hypothesis is speculative, it is clear only the first
layers use sampled positions, so including these in
the input as in Eq. (\ref{eq:inputs}) is sufficient.
\begin{table}[h]
\vspace{-5mm}
\centering
\raisebox{2em}{\small(a)~}
\begin{tabular}[b]{lccc}
\multicolumn{4}{c}{{\bf RoPE Mixing} (S model size)} \\
\hline
 & s/Batch & FID $\downarrow$ & IS $\uparrow$ \\
\hline
No Mixing & 0.181 & 5.32 &234.7  \\
All Layers& 0.195 & 5.33 & 230.2  \\
First 2 Layers & 0.184 & 5.46 & 226.5  \\
\hline
\end{tabular}
\hspace{5em}
\raisebox{4.5em}{\small(b)}
\includegraphics[width=0.2\textwidth]{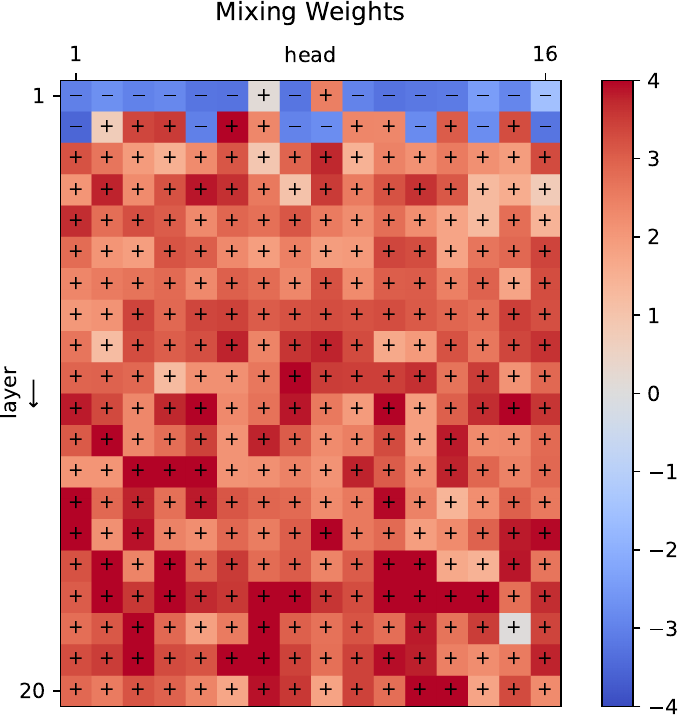} \\
\vspace{-1mm}
\caption{
RoPE mixing strategies (see Sec. \ref{sec:embeddings}).
(a) Results comparison; times use batch size 128 on single GH200.  (b) Mixing
weights when enabled for all layers.  Negative weights (blue)
correspond to sampled positions, while
positive (red) correspond to output locations.
Only the first two layers have negative weights:  
Information for previously-sampled values is extracted early.
}
\vspace{-3mm}
\label{tab:rope-mixing}
\end{table}

\subsection{Entropy Analysis}
\label{sec:entropy}

\begin{figure}[h]
	\centering
	\begin{tabular}{cc}
		\includegraphics[width=0.45\textwidth]{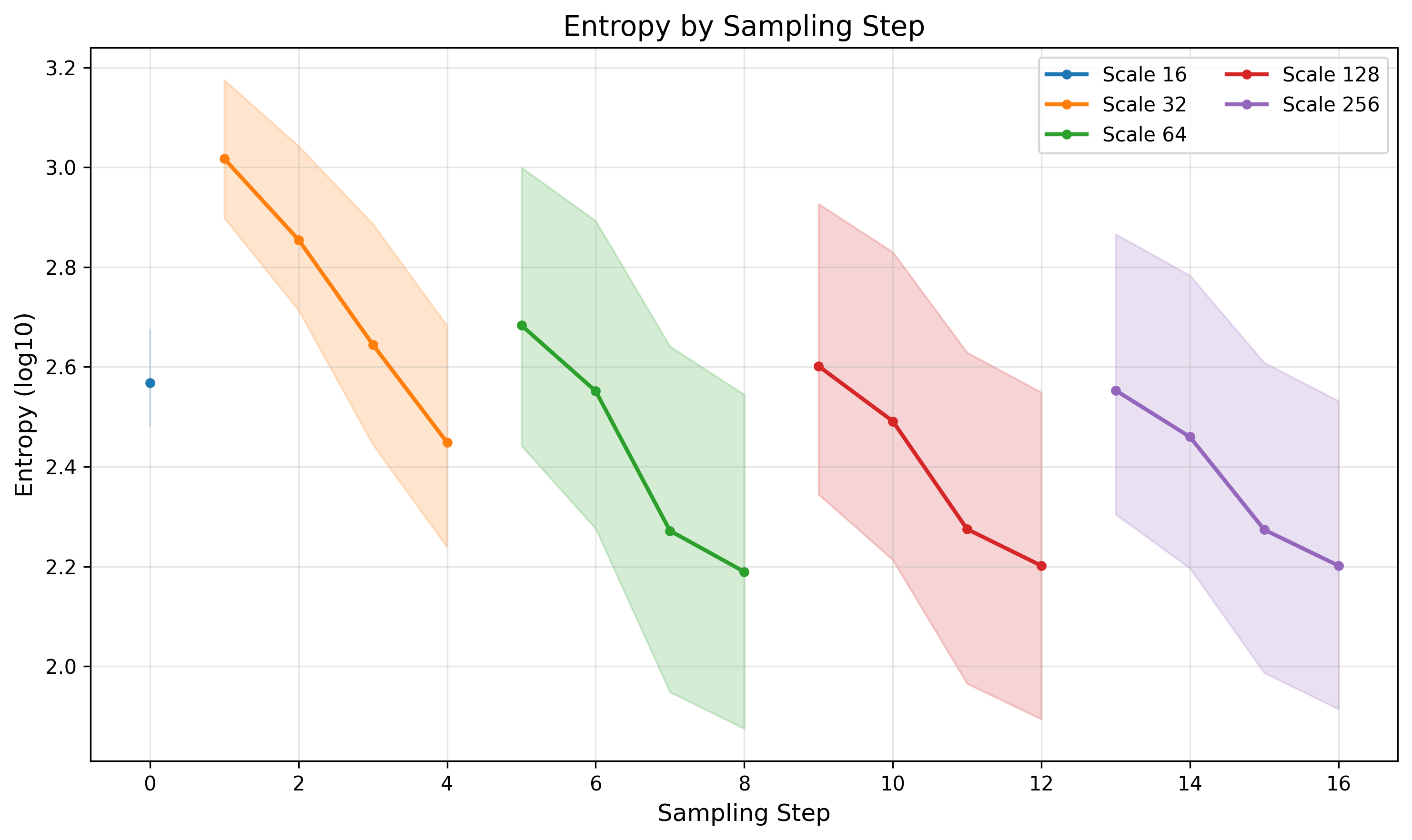} &
		\includegraphics[width=0.45\textwidth]{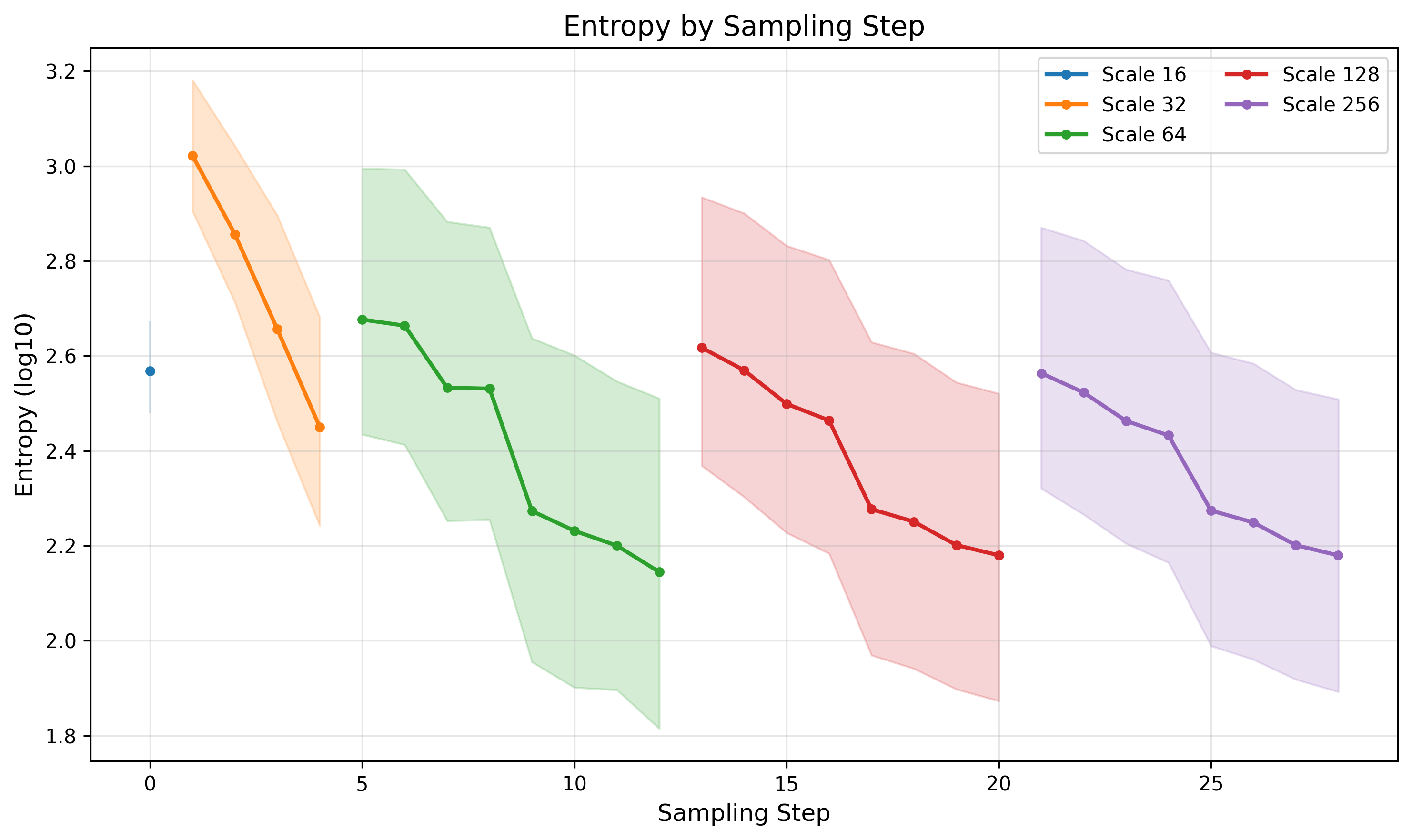} \\[0.5em]
		\hdashline \\[0.0em]
		
		\includegraphics[width=0.45\textwidth]{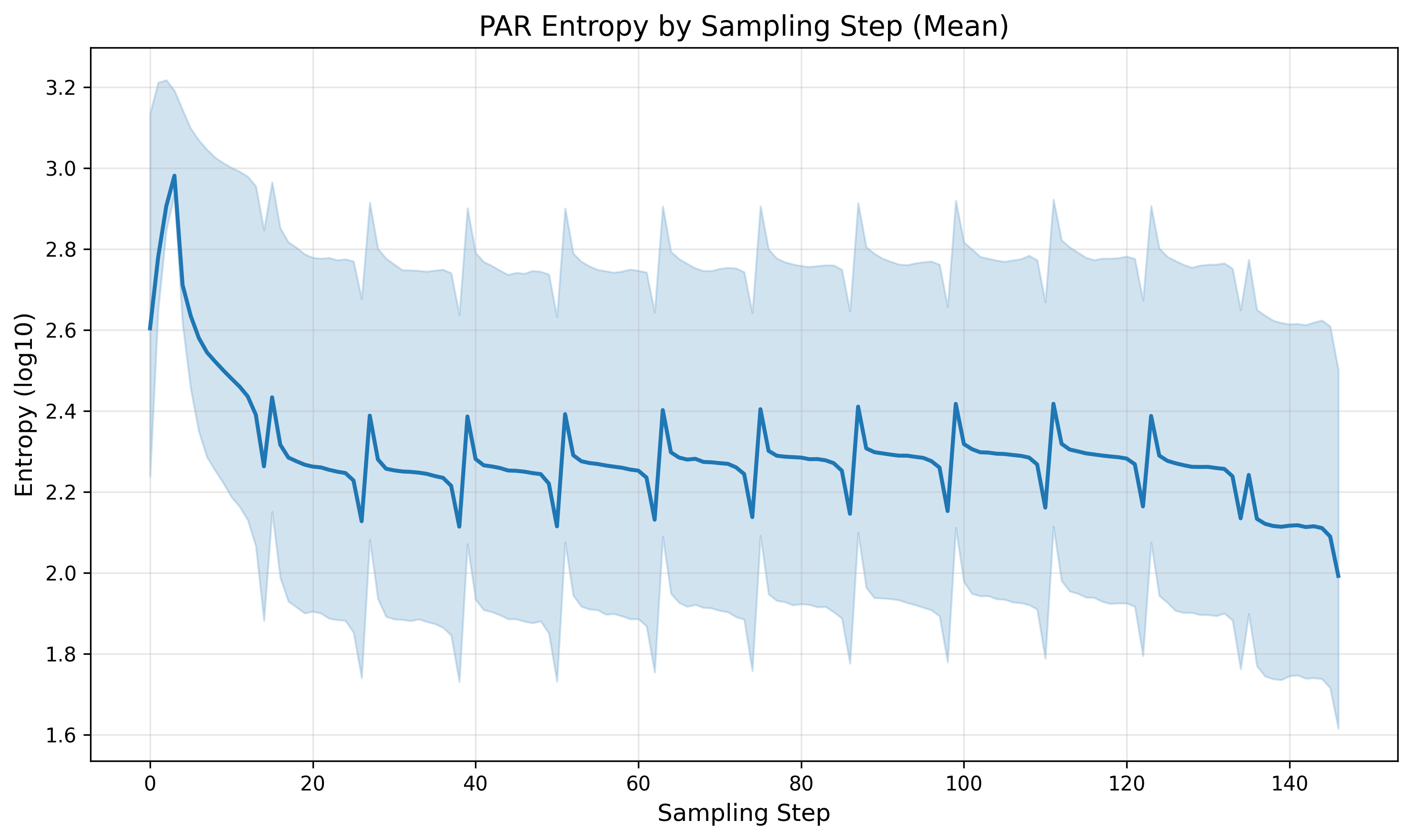} &
		\includegraphics[width=0.45\textwidth]{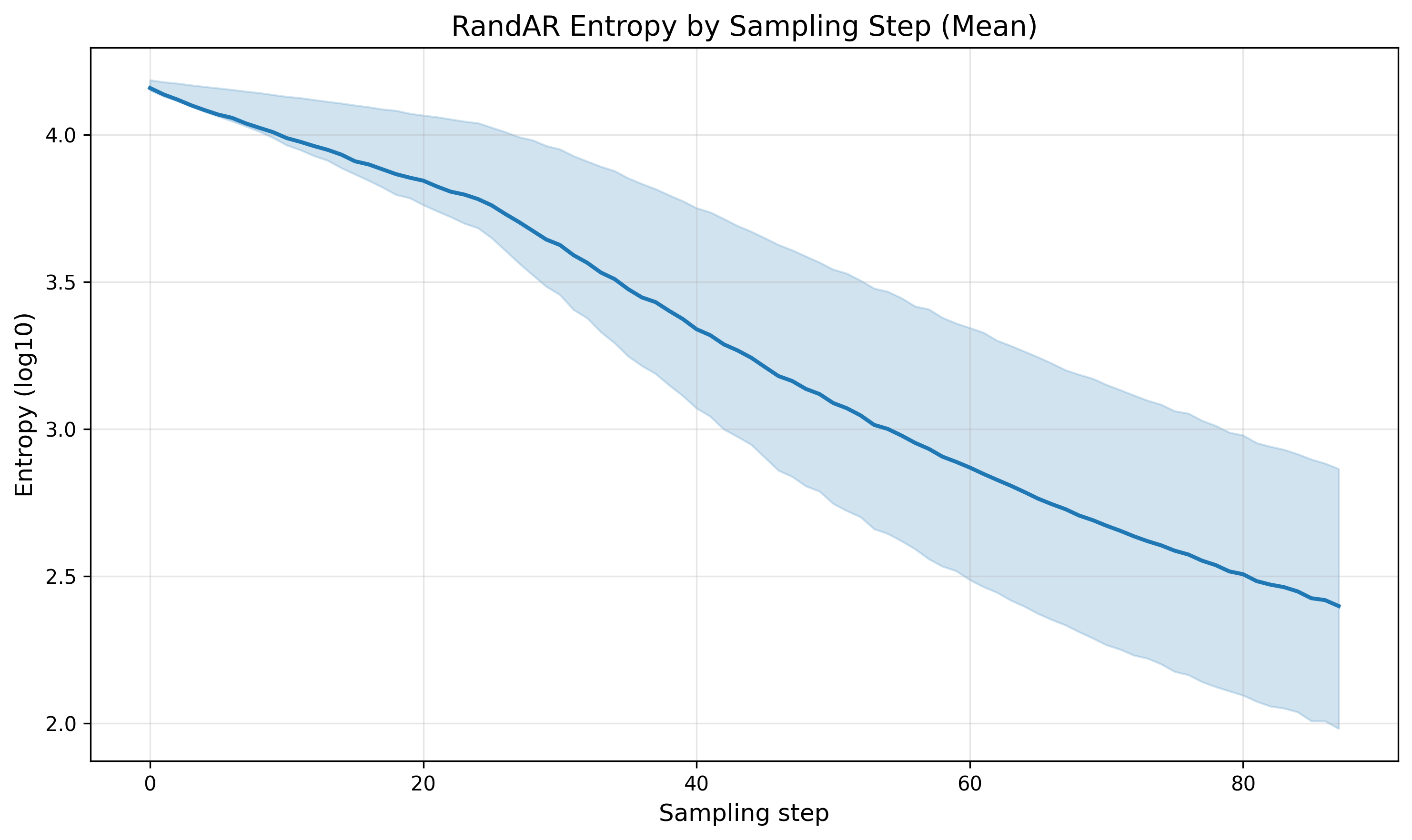}
	\end{tabular}
	\caption{
	Top: Aggregate entropy measurements over 10K samples.
	Entropy decreases within each scale, but
	jumps between scales as new details are introduced.
	Left: 4 steps/scale; Right: 8 steps/scale. \\
	Bottom: Entropy measured for PAR and RandAR, for comparison.
	}
	\label{fig:entropy-aggregate}
\end{figure}

To further illustrate the behavior of multiscale checkerboard sampling, we plot measurements
of entropy for the multinomial distribution at each token during sampling.
Fig. \ref{fig:entropy-aggregate} shows aggregate measurements taken over 10,000
random samples of our L model.  We show mean entropy as well as regions corresponding to
25th and 75th percentiles, over each sampling step.  Appendix \ref{app:entropy-sig-tests}
confirms statistical significance of the following descriptions.
As expected, entropy decreases as steps progress and samples
resolve modes and variance by conditioning.  Between scales, though, there is a jump in
entropy, as additional higher-resolution details are introduced, followed by
continued decrease as sampling within the scale develops.
Interestingly, on average, the largest entropy drop within each scale occurs
half-way through the ordering, when every-other location has been filled.  
This is the transition point where
each unsampled location has 4 adjacent sampled neighbors.

For comparison, we also plot entropy for PAR and RandAR.
PAR samples 4 quadrants in parallel with raster order and
has a sawtooth pattern
corresponding to the raster rows. The last row shows a drop, since the bottom
row of top half is adjacent to already-sampled rows of the bottom half.
RandAR samples in a random order; correspondingly, its entropy
smoothly decreases on average, corresponding to its schedule.

\begin{figure}[h]
	\centering
	\includegraphics[width=0.48\textwidth]{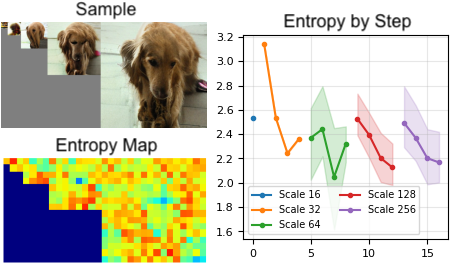}
	\hspace{0.5em}\rule[-0.25ex]{0.4pt}{4cm}\hspace{0.5em}
	\includegraphics[width=0.47\textwidth]{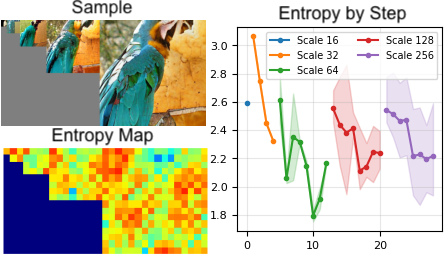}
	\caption{
	Entropy measurements for individual samples.
	Left: 4 steps/scale; Right: 8 steps/scale.
	}
	\label{fig:entropy-indiv}
\end{figure}

Fig. \ref{fig:entropy-indiv} shows entropy for individual samples from our method, both
over steps, and over locations.  Here, we see
a trend consistent with the aggregate measures, but also substantial variation,
since the token distribution depends
heavily on the image content being generated.  Additionally, the entropy maps
clearly display checkerboard patterns, a result of our 
sampling order.  This underscores our method's ability to model local
conditional dependencies, particularly between adjacent locations.


\begin{figure}[h!!!]
\centering
\includegraphics[width=0.95\textwidth]{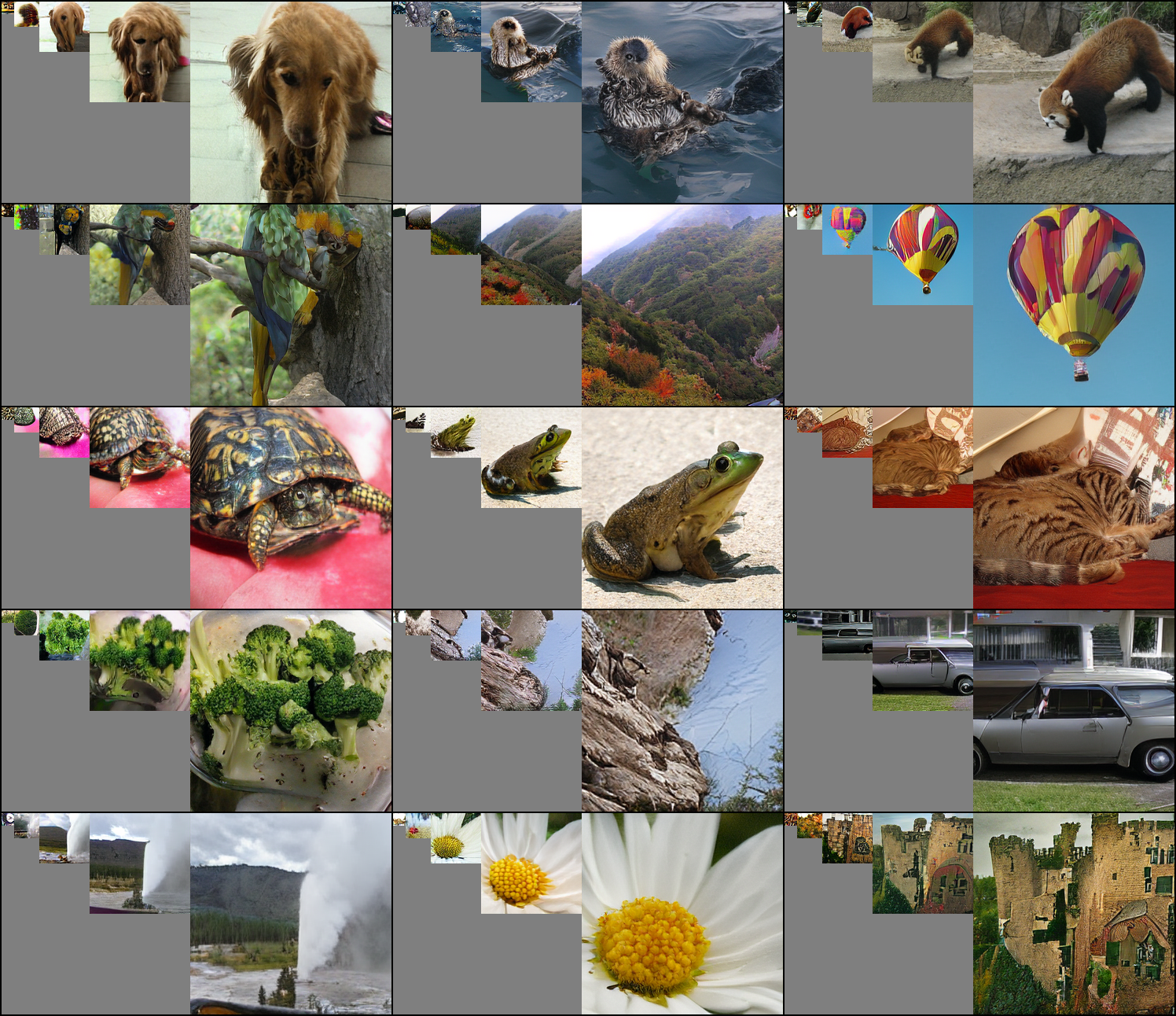}\\
\hspace{0mm}
\includegraphics[width=0.95\textwidth]{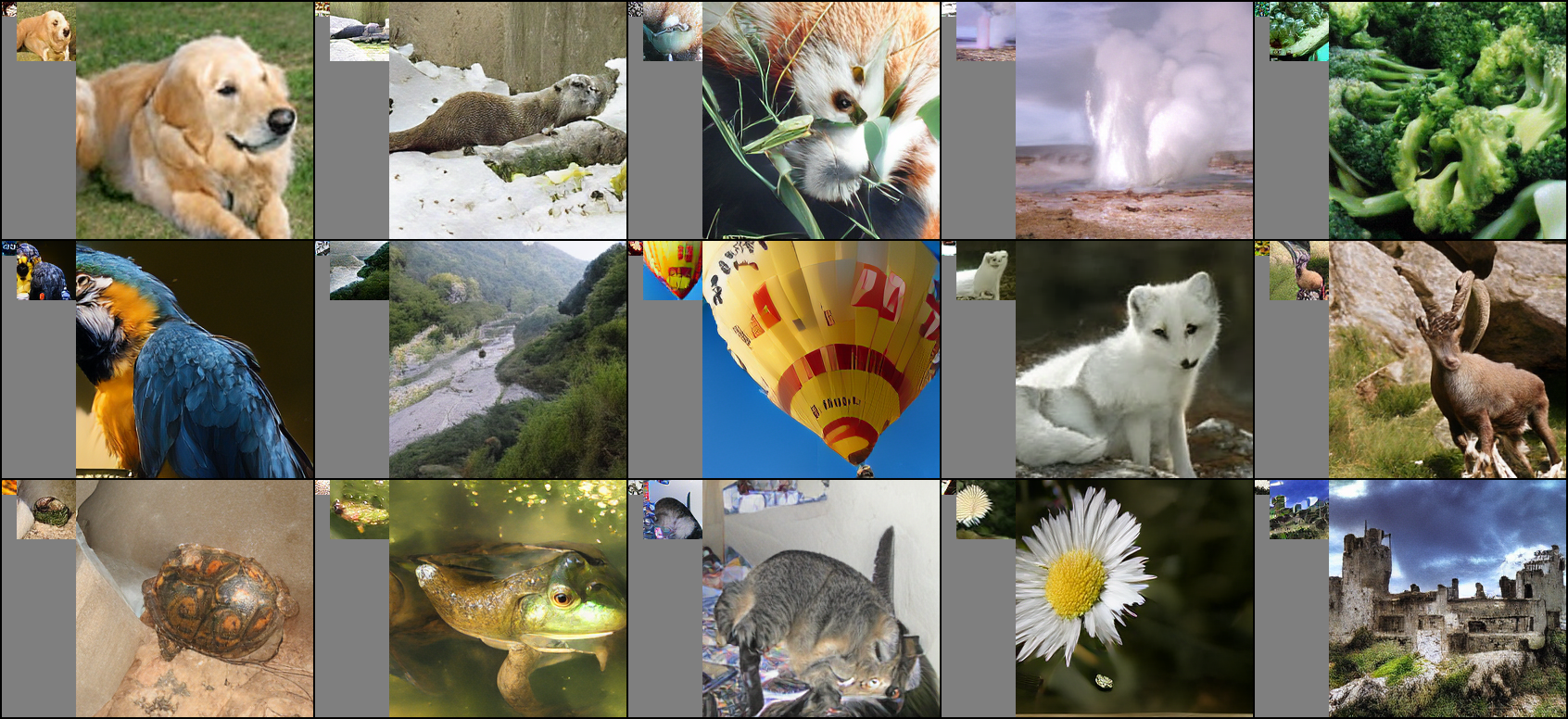}
\caption{
Samples from our Checkerboard-L model, scale factor 2x (top) and 4x (bottom).
}
\label{fig:samples}
\end{figure}

\section{Conclusion}

We have described a multiscale image generation method based on a
spatially-balanced progressive checkerboard sampling order.  By modeling both
between-scale and within-scale conditional dependencies, our method is able to
generate images with fewer sampling steps and competitive inference time
compared to recent methods.  In this setting, our experiments on ImageNet show that
several multiscale pyramid ratios lead to similar performance for a
given number of total sampling steps, enabling our method to use large
scaling ratios of up to 4x.  Future work may include extensions to video or
text modalities, or further studying the relationship between scales and steps
in the context of mutual information and dependency modeling.

\section{Broader Impacts}

Image generation has a wide range of applications, including
creative content generation and data augmentation for
training other models.  It also has potential risks,
including misuse by generating deepfakes or other deceptive
content, surfacing biases in training data and perpetuating
inaccurate representations, including when used for
generating new training data.  Additionally, improved
efficiency, especially in an autoregressive model close to
those used for text generation, may enable widespread
multimodal reasoning models, which could have significant
impacts, such as improved reasoning through visual sketches,
and tighter automatic visual refinement loops; these may
enable new unforeseen applications in turn, with potential
effects both positive and negative.

\bibliography{references}
\bibliographystyle{tmlr}

\newpage

\appendixsection{Experiments at 512x512 Resolution}
\label{app:expers512}

\begin{figure}[h]
	\centering
	
	\includegraphics[width=0.45\textwidth]{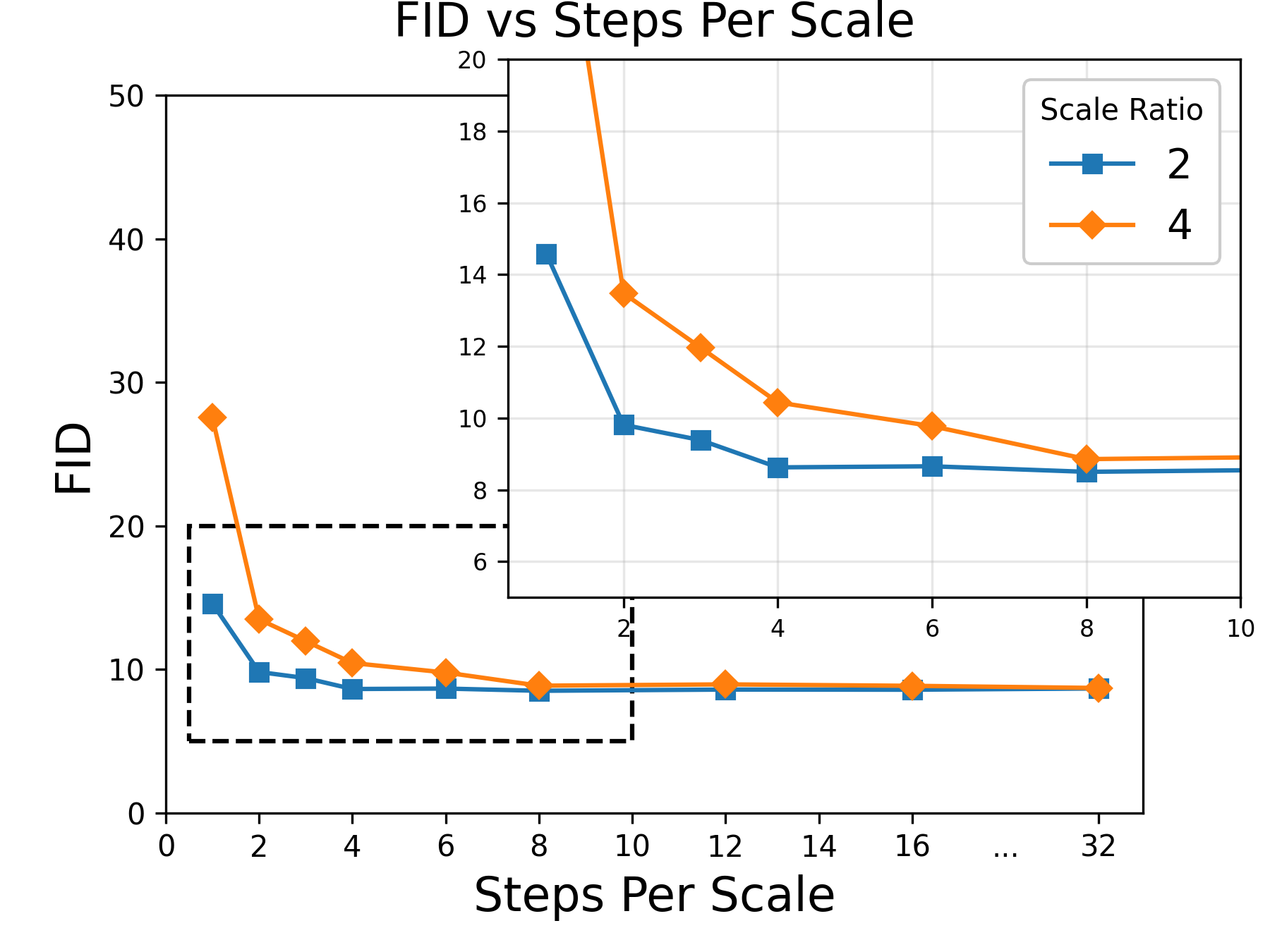}
	\includegraphics[width=0.45\textwidth]{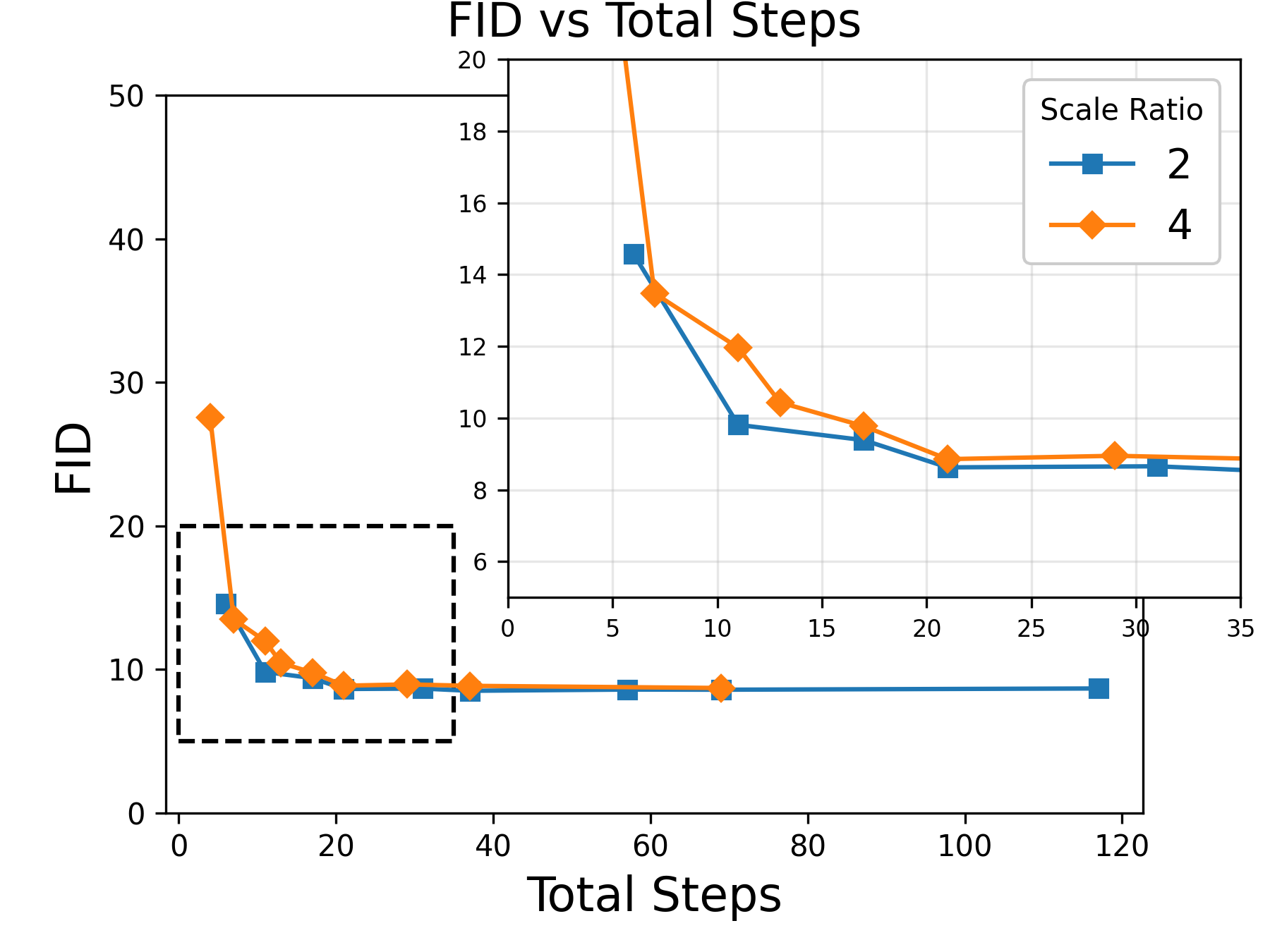}
	\\
	\includegraphics[width=0.44\textwidth]{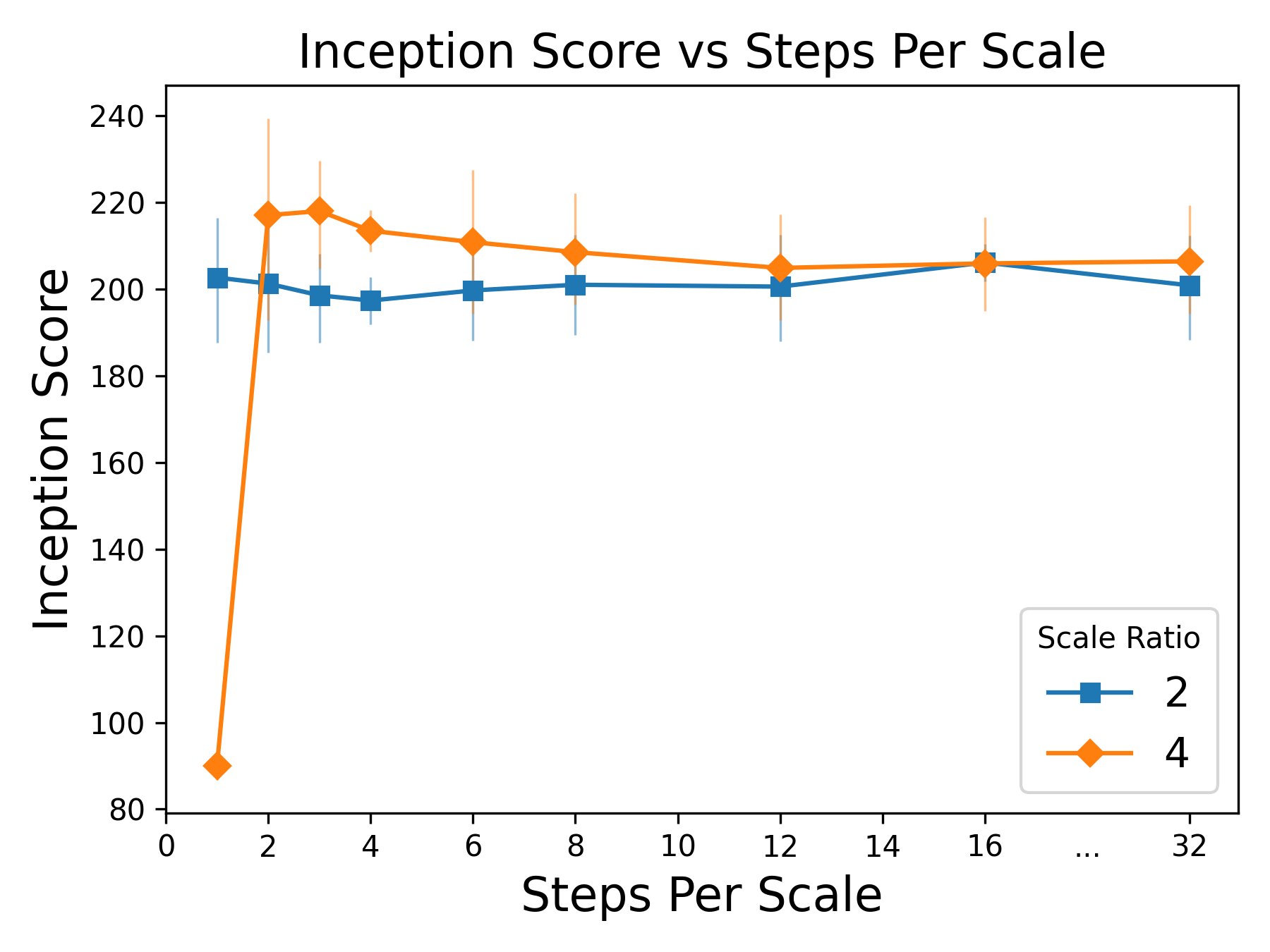}
	\hspace{0.5mm}
	\includegraphics[width=0.44\textwidth]{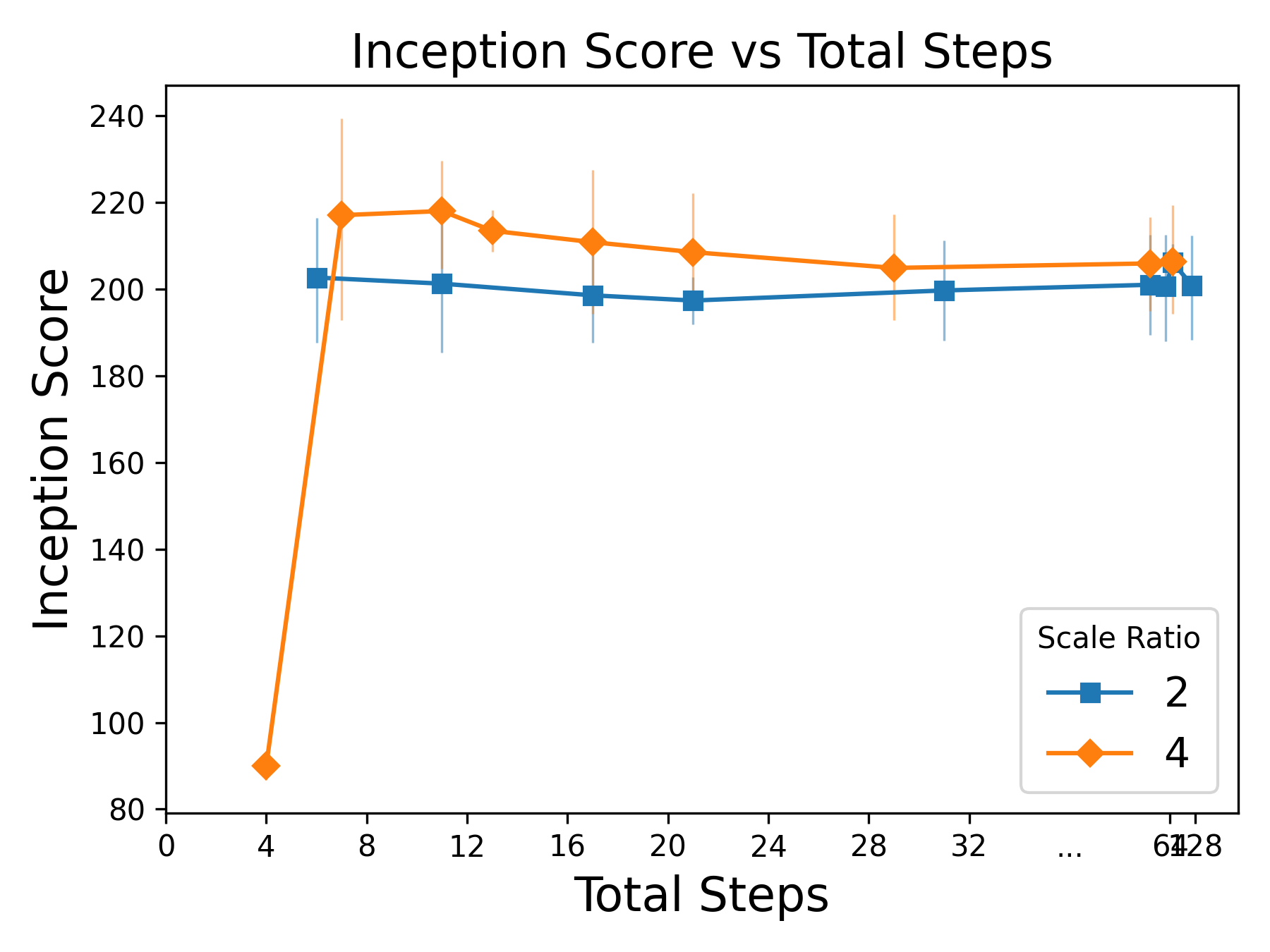}
	\hspace{0.0mm}
	\caption{
	FID and IS by scale ratio and number of steps for the S model, at $512\times512$ resolution
	using incremental transfer learning from the corresponding models trained at 256 resolution (see text for details).
	Left: by numbers of steps per scale; Right: total number of steps.
	}
	\label{fig:scale-ratios-512}
\end{figure}

To investigate whether our findings regarding the relationship between scale
ratio and sampling steps generalize to higher resolutions, we perform a limited
set of experiments at $512\times512$ resolution.

Using the small (S) model size, we train two models at $512\times512$ resolution, using 2x
and 4x scale ratios.  Each are initialized using weights from the corresponding
model trained at 256 resolution, and then trained using a short transfer
schedule for 10 epochs (1 warmup $\rightarrow$ 5 main lr $\rightarrow$ 4
decreased lr with 2 stepdowns).

Note the 2x model simply appends the 32 scale to its pre-existing progression:
$[1,2,4,8,16,32]$.  However, the 4x model has to change all scales in its list, from
$[1,4,16]$ to $[1,2,8,32]$, so is not as natural a transfer setting.

The results are shown in Fig. \ref{fig:scale-ratios-512}, largely supporting
our findings from the 256 resolution experiments.
Viewed by total
steps, the models' FID curves are close to one another, both reaching best FID
at 21 total steps.  In this case, the 2x model is slightly better in FID than
the 4x model.  However, it's unclear if this is due to fundamental differences
or experimental
limitations, since the 4x model's scale progression is less natural in this
transfer setting.
Overall, the results suggest that while the exact scale ratio may start to have some more impact on
performance at 512 resolution, the total number of steps is still a dominant factor, and that
multiple scale ratios can achieve similar performance when using the same total
number of steps.

\newpage
\appendixsection{Total Steps Evaluations for L Model}
\label{app:scale-steps-L256}

\begin{figure}[h]
	\centering
	
	\includegraphics[width=0.45\textwidth]{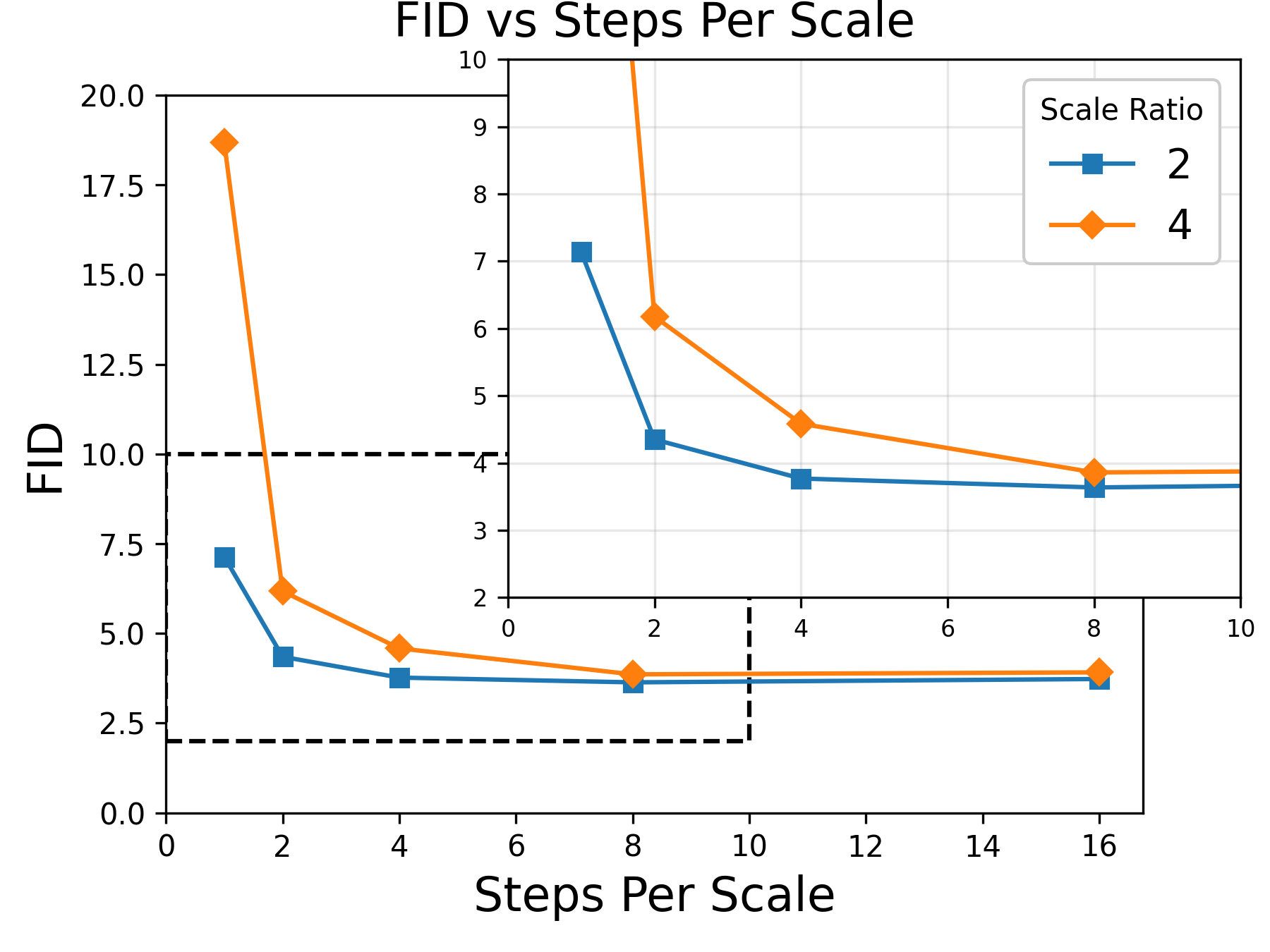}
	\includegraphics[width=0.45\textwidth]{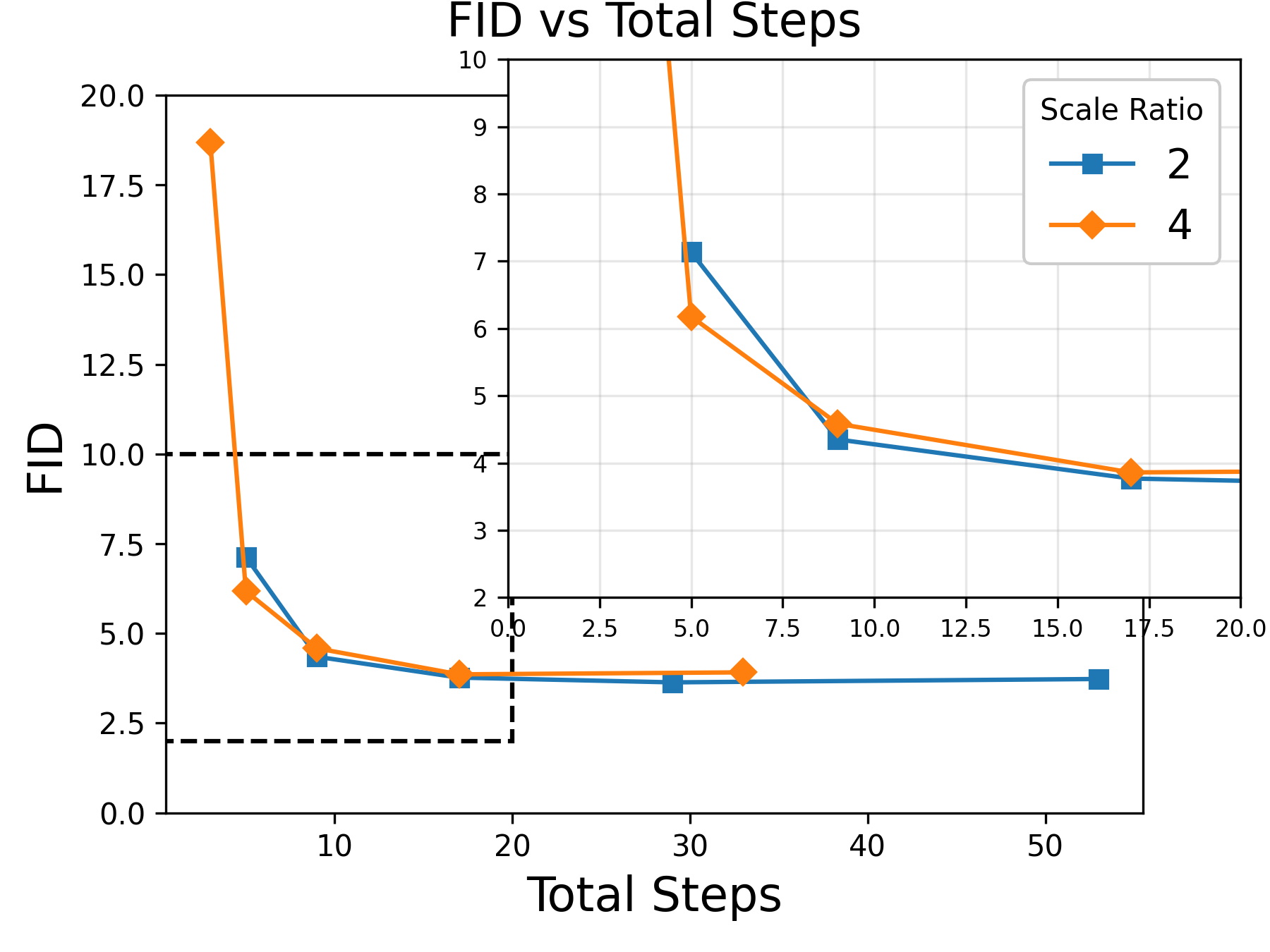}
	\\
	\includegraphics[width=0.44\textwidth]{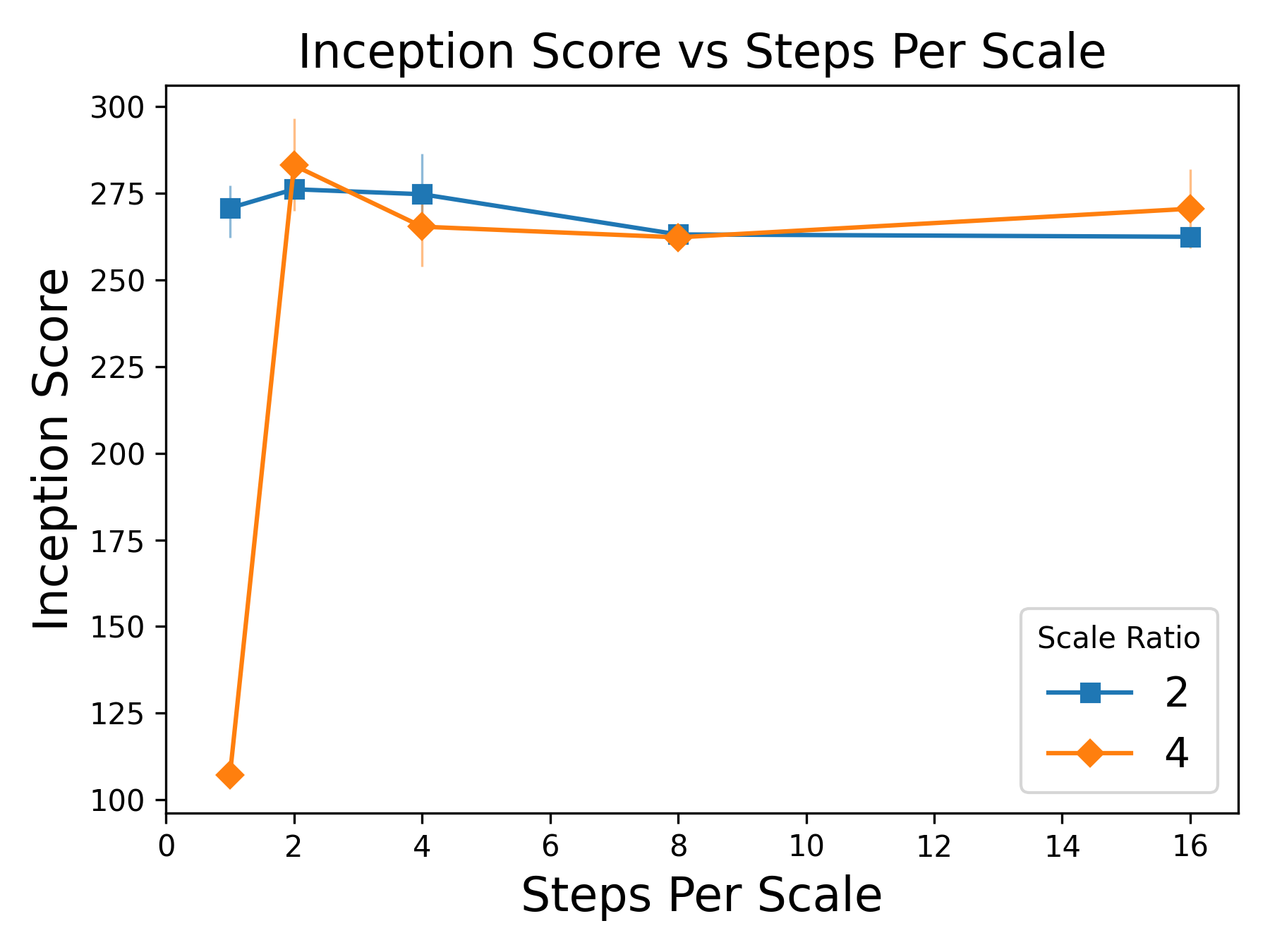}
	\hspace{0.5mm}
	\includegraphics[width=0.44\textwidth]{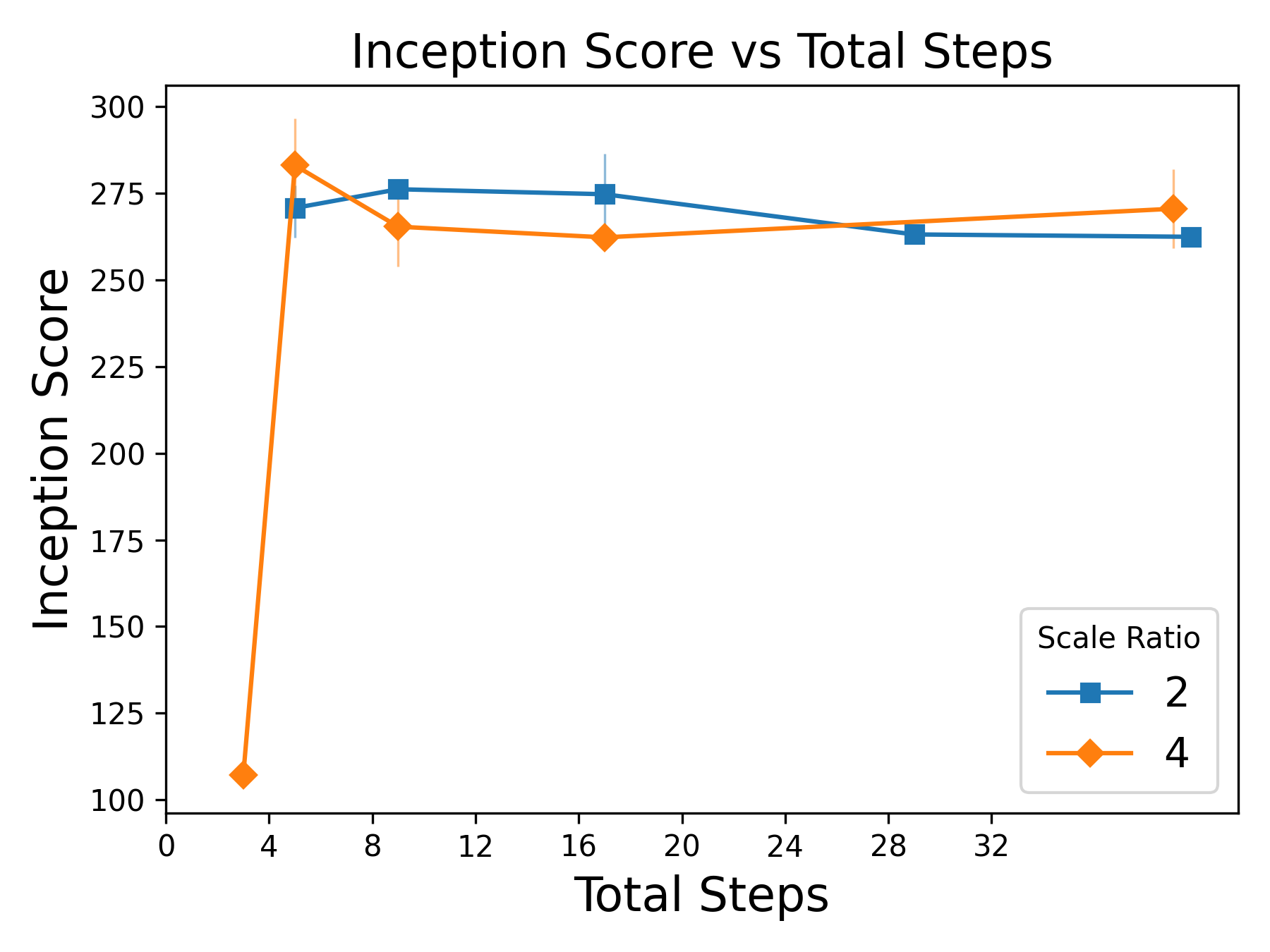}
	\hspace{0.0mm}
	\caption{
	FID and IS by scale ratio and number of steps for the L model (256 resolution).
	}
	\label{fig:scale-ratios-L256}
\end{figure}

Using the large (L) model, we plot FID and IS by scale ratio and number of
steps in Fig. \ref{fig:scale-ratios-L256}.

The results are consistent with those from the S model, showing that while 4x
scaling requires more steps per scale than 2x, when viewed by total steps,
performance is similar between the two.  While the 2x ratio may achieve
slightly better FID, the total number of steps is still dominant, with best FID
achieved at 17 total steps for both ratios.

\newpage
\appendixsection{Qualitative Failure Cases}
\label{app:failurecases}

Below we show some qualitative failure cases from our Checkerboard-L model at
256 resolution, using 4 steps per scale.  Although none can be linked
conclusively to artifacts from checkerboard sampling, cases of medium- and
long-range object distortion, particularly of highly geometric objects like
circles and straight lines, might be influenced by this if token samples are
out of place from their long-range alignment.

\begin{figure}[h]
\centering
\includegraphics[width=0.95\textwidth]{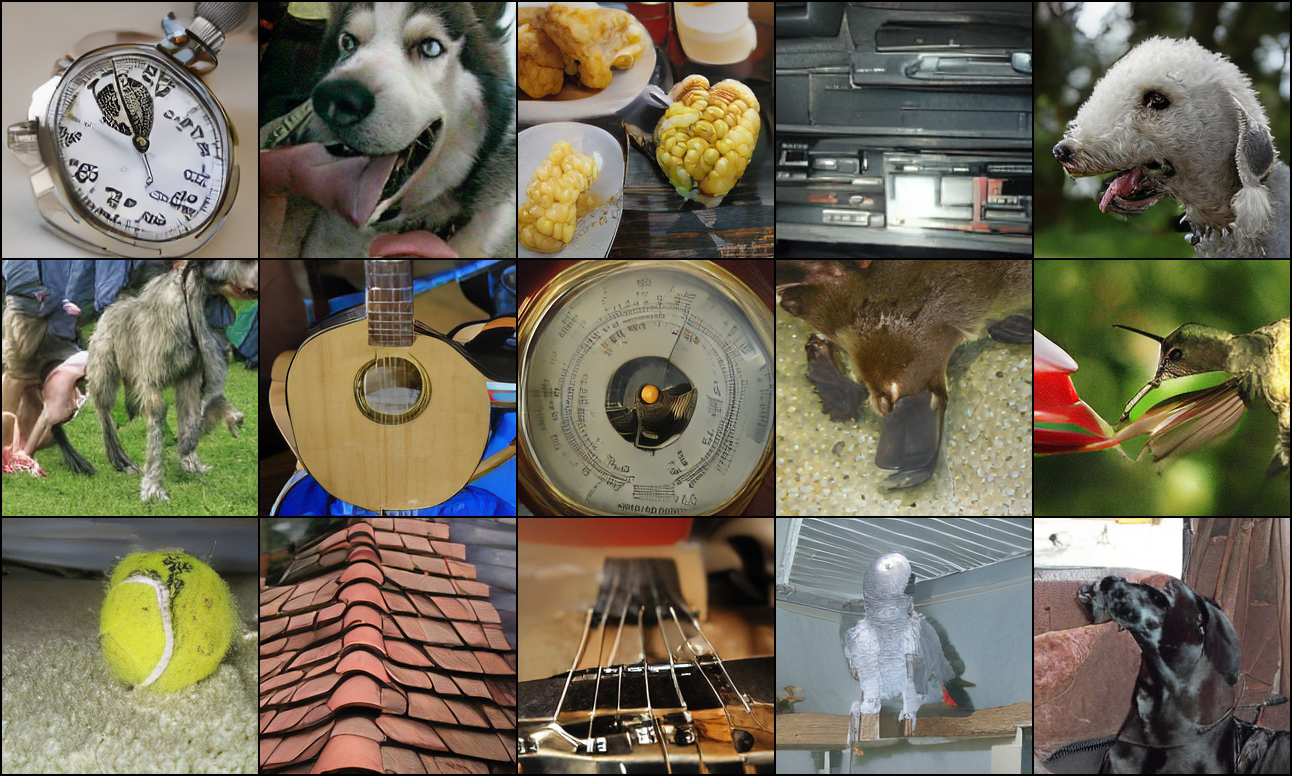}\\
\caption{
Qualitative failure cases from our Checkerboard-L model, scale factor 2x.
}
\label{fig:failurecases}
\end{figure}

\newpage
\appendixsection{Entropy Analysis Statistics}
\label{app:entropy-sig-tests}

As described in Sec. \ref{sec:entropy}, entropy
decreases within each scale on average, with the average largest decrease half-way through
the ordering, when every-other location has been filled.  Here we confirm
the statistical significance of these observations.

First, we check the entropy decrease for each sampling step.
For $P=4$ steps per scale, entropy is on average lower in
every step $t$ compared to step $t-1$ ($t$ $\le -28.2$; $p < 10^{-100}$).  This
is also the case for the 8x8 and 16x16 scales with $P=8$ ($t \le -11.7$; $p < 10^{-30}$).
For the 4x4 scale with $P=8$, only two locations are sampled per step, and
the decrease is significant at all but two steps, visible as flat lines
for this scale in Fig \ref{fig:entropy-aggregate} (right side, green line).
Full results are in Table \ref{tab:entropy-testa} below.

Next, we check the entropy drop at the half-way point is the largest drop as
described, comparing its size compared to the next-largest drop in each scale.
Both $P=4$ and $P=8$ steps per scale have signficantly larger decreases at
their half-way points ($t \le -24.1$; $p < 10^{-100}$).  Full results are
in Table \ref{tab:entropy-testb}.

\begin{table}[h]
\centering
\tiny
\begin{tabular}{c | l rrrr | l rrrr}
\toprule
 & \multicolumn{5}{c|}{$P=4$} & \multicolumn{5}{c}{$P=8$} \\
\cmidrule(r){2-6}\cmidrule(l){7-11}
Scale & $t$ & Mean $\Delta H$ & SE & SD & $t$-stat & $t$ & Mean $\Delta H$ & SE & SD & $t$-stat \\
\midrule
\multirow{7}{*}{$4\!\times\!4^\dagger$}
  & $1$   & $-0.315$ & $0.006$ & $0.607$ & $\ \ -51.9$  & $1$   & $-0.017$ & $0.007$ & $0.713$ & $\ \ \ -2.4^\ddagger$ \\
  & $2^*$ & $-0.661$ & $0.006$ & $0.597$ & $-110.9$     & $2$   & $-0.324$ & $0.008$ & $0.766$ & $\ \ -42.3$ \\
  & $3$   & $-0.182$ & $0.006$ & $0.644$ & $\ \ -28.2$  & $3$   & $-0.009$ & $0.007$ & $0.748$ & $\ \ \ -1.2^\ddagger$ \\
  &       & \multicolumn{4}{c|}{---}           & $4^*$ & $-0.594$ & $0.008$ & $0.788$ & $\ \ -75.4$ \\
  &       & \multicolumn{4}{c|}{---}           & $5$   & $-0.111$ & $0.008$ & $0.812$ & $\ \ -13.7$ \\
  &       & \multicolumn{4}{c|}{---}           & $6$   & $-0.063$ & $0.008$ & $0.838$ & $\ \ \ -7.5$ \\
  &       & \multicolumn{4}{c|}{---}           & $7$   & $-0.130$ & $0.008$ & $0.813$ & $\ \ -16.0$ \\
\midrule
\multirow{7}{*}{$8\!\times\!8$}
  & $1$   & $-0.256$ & $0.003$ & $0.308$ & $\ \ -83.0$  & $1$   & $-0.116$ & $0.003$ & $0.342$ & $\ \ -34.0$ \\
  & $2^*$ & $-0.512$ & $0.003$ & $0.297$ & $-172.2$     & $2$   & $-0.170$ & $0.004$ & $0.372$ & $\ \ -45.7$ \\
  & $3$   & $-0.167$ & $0.003$ & $0.318$ & $\ \ -52.7$  & $3$   & $-0.076$ & $0.004$ & $0.351$ & $\ \ -21.6$ \\
  &       & \multicolumn{4}{c|}{---}           & $4^*$ & $-0.439$ & $0.004$ & $0.386$ & $-113.8$ \\
  &       & \multicolumn{4}{c|}{---}           & $5$   & $-0.053$ & $0.004$ & $0.377$ & $\ \ -14.0$ \\
  &       & \multicolumn{4}{c|}{---}           & $6$   & $-0.118$ & $0.004$ & $0.405$ & $\ \ -29.2$ \\
  &       & \multicolumn{4}{c|}{---}           & $7$   & $-0.044$ & $0.004$ & $0.375$ & $\ \ -11.7$ \\
\midrule
\multirow{7}{*}{$16\!\times\!16$}
  & $1$   & $-0.214$ & $0.001$ & $0.148$ & $-144.4$     & $1$   & $-0.094$ & $0.002$ & $0.157$ & $\ \ -60.0$ \\
  & $2^*$ & $-0.435$ & $0.001$ & $0.147$ & $-296.3$     & $2$   & $-0.141$ & $0.002$ & $0.179$ & $\ \ -78.7$ \\
  & $3$   & $-0.163$ & $0.002$ & $0.151$ & $-107.9$     & $3$   & $-0.071$ & $0.002$ & $0.161$ & $\ \ -44.1$ \\
  &       & \multicolumn{4}{c|}{---}           & $4^*$ & $-0.370$ & $0.002$ & $0.184$ & $-200.9$ \\
  &       & \multicolumn{4}{c|}{---}           & $5$   & $-0.054$ & $0.002$ & $0.172$ & $\ \ -31.5$ \\
  &       & \multicolumn{4}{c|}{---}           & $6$   & $-0.112$ & $0.002$ & $0.193$ & $\ \ -57.9$ \\
  &       & \multicolumn{4}{c|}{---}           & $7$   & $-0.044$ & $0.002$ & $0.173$ & $\ \ -25.3$ \\
\bottomrule
\end{tabular}
\caption{
\small
  Mean adjacent-step entropy change $\Delta H$ (nats),
  for $P=4$ and $P=8$ steps/scale ($N=10{,}000$ samples, CFG=1.5).
  $^*$marks the midpoint step.
  All $t$-statistics correspond to $p < 10^{-10}$ (one-tailed $t$-test),
  except where marked $^\ddagger$ (not significant).
}
\label{tab:entropy-testa}
\end{table}

\begin{table}[h]
\centering
\tiny
\begin{tabular}{c | l rrrr | l rrrr}
\toprule
 & \multicolumn{5}{c|}{$P=4$} & \multicolumn{5}{c}{$P=8$} \\
\cmidrule(r){2-6}\cmidrule(l){7-11}
Scale & vs.\ $t$ & Mean $\delta$ & SE & SD & $t$-stat & vs.\ $t$ & Mean $\delta$ & SE & SD & $t$-stat \\
\midrule
\multirow{7}{*}{$4\!\times\!4$}
  & $1$   & $-0.346$ & $0.010$ & $1.030$ & $-33.6$ & $1$   & $-0.577$ & $0.011$ & $1.080$ & $-53.5$ \\
  & $2^*$ & $0$      & ---     & ---     & ---     & $2$   & $-0.270$ & $0.011$ & $1.119$ & $-24.2$ \\
  & $3$   & $-0.480$ & $0.011$ & $1.089$ & $-44.1$ & $3$   & $-0.585$ & $0.013$ & $1.319$ & $-44.4$ \\
  &       & \multicolumn{4}{c|}{---}      & $4^*$ & $0$      & ---     & ---     & ---     \\
  &       & \multicolumn{4}{c|}{---}      & $5$   & $-0.483$ & $0.014$ & $1.396$ & $-34.6$ \\
  &       & \multicolumn{4}{c|}{---}      & $6$   & $-0.531$ & $0.012$ & $1.175$ & $-45.2$ \\
  &       & \multicolumn{4}{c|}{---}      & $7$   & $-0.464$ & $0.012$ & $1.147$ & $-40.4$ \\
\midrule
\multirow{7}{*}{$8\!\times\!8$}
  & $1$   & $-0.256$ & $0.005$ & $0.522$ & $-49.1$ & $1$   & $-0.323$ & $0.005$ & $0.523$ & $-61.7$ \\
  & $2^*$ & $0$      & ---     & ---     & ---     & $2$   & $-0.269$ & $0.006$ & $0.559$ & $-48.1$ \\
  & $3$   & $-0.344$ & $0.005$ & $0.543$ & $-63.3$ & $3$   & $-0.363$ & $0.006$ & $0.631$ & $-57.5$ \\
  &       & \multicolumn{4}{c|}{---}      & $4^*$ & $0$      & ---     & ---     & ---     \\
  &       & \multicolumn{4}{c|}{---}      & $5$   & $-0.386$ & $0.007$ & $0.658$ & $-58.7$ \\
  &       & \multicolumn{4}{c|}{---}      & $6$   & $-0.321$ & $0.006$ & $0.582$ & $-55.1$ \\
  &       & \multicolumn{4}{c|}{---}      & $7$   & $-0.395$ & $0.006$ & $0.546$ & $-72.3$ \\
\midrule
\multirow{7}{*}{$16\!\times\!16$}
  & $1$   & $-0.221$ & $0.003$ & $0.248$ & $-89.4$  & $1$   & $-0.276$ & $0.002$ & $0.239$ & $-115.4$ \\
  & $2^*$ & $0$      & ---     & ---     & ---      & $2$   & $-0.230$ & $0.003$ & $0.272$ & $-84.3$  \\
  & $3$   & $-0.272$ & $0.003$ & $0.257$ & $-106.2$ & $3$   & $-0.299$ & $0.003$ & $0.290$ & $-103.1$ \\
  &       & \multicolumn{4}{c|}{---}       & $4^*$ & $0$      & ---     & ---     & ---      \\
  &       & \multicolumn{4}{c|}{---}       & $5$   & $-0.316$ & $0.003$ & $0.303$ & $-104.3$ \\
  &       & \multicolumn{4}{c|}{---}       & $6$   & $-0.259$ & $0.003$ & $0.283$ & $-91.5$  \\
  &       & \multicolumn{4}{c|}{---}       & $7$   & $-0.327$ & $0.003$ & $0.252$ & $-129.4$ \\
\bottomrule
\end{tabular}
\caption{
\small
  Difference between half-way step and every other step, $\delta = \Delta H_{\mathrm{mid}} - \Delta H_t$ (nats).
  $\delta < 0$ at every comparison confirms the middle step has the largest
  entropy change (one-tailed $t$-test), 
  all $t$-statistics are significant with $p < 10^{-100}$.
}
\label{tab:entropy-testb}
\end{table}

\newpage
\appendixsection{Further Discussion on Relationship Between Scale Ratio and Sampling Steps}
\label{app:scale-steps-mechanism}

In this section, we provide additional discussion on possible reasons why the
scale-steps relationship shown in Sec. \ref{sec:scale-steps-evals} arises.

Consider the following $4\times4$ area, with four of the 16 locations labeled for
reference:

\[
\begin{matrix}
A	  & \cdot & B     & \cdot \\
\cdot & \cdot & \cdot & \cdot \\
C     & \cdot & D     & \cdot \\
\cdot & \cdot & \cdot & \cdot
\end{matrix}
\]

The question here is, how many locations can we sample independently?

First, consider a 2x scaling ratio.  When scaling up, each pixel at scale $i-1$
"turns into" a $2\times2$ window of pixels in scale $i$.  In the example above, this
means that we've already sampled four values in scale $i-1$, corresponding to the
$2\times2$ locations anchored at A, B, C and D.  So for the most part,
A, B, C and D are conditionally independent of each other, given samples at
the corresponding locations in the previous scale.
In practice, this is not entirely true --- there could be fine-grained textures
that need to be consistent between A and B in the new scale, for example --- but it's
mostly the case.  Thus we can sample all four at the same time.

In general, this suggests that when upsampling with a ratio S, so that each
pixel at scale $i-1$ ``turns into'' a window of $S\times S$ pixels at scale $i$, we can
sample each window in parallel, but the $S\times S$ locations within each window may still
need to be sampled sequentially.

For a $S=2$ ratio, corresponding to $S\times S=4$, this would imply 4 steps/scale, which
is what we see in practice.

For a $S=4$ ratio, corresponding to $S\times S=16$, this would imply 16 steps/scale.
However, as shown by our measurements in Sec. \ref{sec:scale-steps-evals}, we
can use only 8 and get similar results.  That is, A and D in the above grid can
be sampled at the same time, even though they are within the same $S\times S$ window.
Thus, it's possible to view the 8 steps/scale as an approximation of the 16 we
know are sufficient according to the conditional independence induced by
upsampling.

A key enabler here is very likely information redundancy: that is, information
provided by some samples may overlap with the information provided by other
samples.  In particular, the information provided from samples of A,B,C,D above
using $S=4$x upsampling overlaps information that would have been supplied in the
previous scale using $S=2$x upsampling.  This is because of the balanced sampling
order, which spaces out samples uniformly in the image area.  Once A, B, C and
D are sampled, we are already conditioning on some information from each $2\times2$
window, even when scaling up by 4x:  part of the information that would have
been provided in each $2\times2$ window is also part of the A, B, C and D sampled
values.  The redundancy in the information enables more aggressive parallel
sampling than might be predicted by the sequential-within-each-window
assumptions --- how much more parallelism is enabled by this mechanism depends
on the data.

This information redundancy provides a basis for starting to understand why
different scale ratios achieve similar quality at the same total step count.
The balanced sampling order partially compensates for larger upsampling ratios,
making the system insensitive to the exact ratio used, so that within a range
of scale factors, performance is largely determined by the total number of
steps.

\end{document}